\documentclass[sigconf]{aamas}  

\AtBeginDocument{%
  \providecommand\BibTeX{{%
    \normalfont B\kern-0.5em{\scshape i\kern-0.25em b}\kern-0.8em\TeX}}}
    
\usepackage{booktabs}    
\usepackage{flushend} 
\usepackage{mysymbols}
\usepackage{algorithm}
\usepackage{algpseudocode}

\usepackage{amsmath,amssymb, amsthm}
\usepackage{bm}
\usepackage{booktabs}
\usepackage{caption}
\usepackage{cleveref}
\usepackage{enumitem}
\usepackage{graphicx}
\usepackage{latexsym}
\usepackage{multicol}
\usepackage{stmaryrd}
\usepackage{subcaption}
\usepackage[T1]{fontenc}

\DeclareMathOperator*{\argmax}{argmax}

\definecolor{bettergreen}{HTML}{4faa42}

\setcopyright{ifaamas}  
\copyrightyear{2020} 
\acmYear{2020} 
\acmDOI{} 
\acmPrice{} 
\acmISBN{} 
\acmConference[AAMAS'20]{Proc.\@ of the 19th International Conference on Autonomous Agents and Multiagent Systems (AAMAS 2020)}{May 9--13, 2020}{Auckland, New Zealand}{B.~An, N.~Yorke-Smith, A.~El~Fallah~Seghrouchni, G.~Sukthankar (eds.)}  

\begin{document}

\title{Hierarchical Cooperative Multi-Agent Reinforcement Learning with Skill Discovery}  


\author{Jiachen Yang}
\authornote{Work done at EA Digital Platform, Data \& AI, Electronic Arts.}
\affiliation{%
  \institution{Georgia Institute of Technology}
  \city{Atlanta} 
  \state{Georgia} 
  \postcode{30332}
}
\email{jiachen.yang@gatech.edu}
\author{Igor Borovikov}
\affiliation{%
  \institution{Electronic Arts}
  \city{Redwood City} 
  \state{California} 
  \postcode{94065}
}
\email{iborovikov@ea.com}
\author{Hongyuan Zha}
\affiliation{%
  \institution{Georgia Institute of Technology}
  \city{Atlanta} 
  \state{Georgia} 
  \postcode{30332}
}
\email{zha@cc.gatech.edu}


\begin{abstract}
Human players in professional team sports achieve high level coordination by dynamically choosing complementary skills and executing primitive actions to perform these skills.
As a step toward creating intelligent agents with this capability for fully cooperative multi-agent settings, we propose a two-level hierarchical multi-agent reinforcement learning (MARL) algorithm with unsupervised skill discovery.
Agents learn useful and distinct skills at the low level via independent Q-learning, while they learn to select complementary latent skill variables at the high level via centralized multi-agent training with an extrinsic team reward.
The set of low-level skills emerges from an intrinsic reward that solely promotes the decodability of latent skill variables from the trajectory of a low-level skill, without the need for hand-crafted rewards for each skill.
For scalable decentralized execution, each agent independently chooses latent skill variables and primitive actions based on local observations.
Our overall method enables the use of general cooperative MARL algorithms for training high level policies and single-agent RL for training low level skills.
Experiments on a stochastic high dimensional team game show the emergence of useful skills and cooperative team play.
The interpretability of the learned skills show the promise of the proposed method for achieving human-AI cooperation in team sports games.
\end{abstract}

%


\maketitle


\section{Introduction}
\label{sec:introduction}

Fully cooperative multi-agent reinforcement learning (MARL) is an active area of research \citep{panait2005cooperative,hernandez2018multiagent} with a diverse set of real-world application, which include autonomous navigation \citep{cao2013overview}, game AI micromanagement \citep{foerster2018counterfactual,rashid2018a}, and traffic network optimization \cite{zhang2019integrating}.
A unique challenge is the need for centralized training for agents to find global optimal cooperative policies, while ensuring scalable decentralized execution whereby agents choose actions independently.
In this paradigm of centralized training with decentralized execution \citep{bernstein2002complexity}, a common approach \citep{sunehag2018value,foerster2018counterfactual,rashid2018a,yang2018cm3,son2019qtran} is to conduct centralized training at the level of \textit{primitive} actions, which are the actions used in the transition function of the Markov game \citep{littman1994markov}.
However, the design of hierarchical agents who can cooperate at a higher level of abstraction using temporally-extended \textit{skills} in high-dimensional multi-agent environments is still an open question.
A skill is a policy that is conditioned on a latent variable, executed for an extended duration, and generates behavior from which the latent variable can be decoded \citep{eysenbach2018diversity,achiam2018variational}.
It is also not clear how multiple agents can \textit{discover} skills without hand-crafted reward functions for each skill, and how to construct such hierarchical policies to allow human interpretation of skills for potential human-AI cooperation.

In this paper, we take a hierarchical approach to fully cooperative MARL and address these questions by drawing inspiration from team sports games.
At the team level, coaches train human players to execute complementary skills in parallel, such as moving to different field positions in a formation, as well as effective sequences of skills over time, such as switching between offensive and defensive maneuvers when ball possession changes.
At the individual level, each player learns a sequence of primitive actions to execute a chosen skill.
Hierarchical approaches inspired from such real-world practices have several benefits for fully cooperative MARL.
From an algorithmic viewpoint, a hierarchical decomposition in two key dimensions---over agents, and across time---simultaneously addresses both the difficulty of learning cooperation at the level of noisy low-level actions in stochastic environments and the difficulty of long-term credit assignment due to highly-delayed rewards (e.g., scoring a goal in football) \citep{ghavamzadeh2006hierarchical,vezhnevets2017feudal}.
Hierarchical approaches may also reduce computational complexity \citep{sutton1999between} to address the exponential increase in sample complexity with number of agents in MARL.
From the viewpoint of human-AI cooperation, which has near-term application to video game AI to improve human players' experiences \citep{zhao2019winning}, hierarchical policies trained with explicit skills is a key step toward interpretable and modular policies.
In this work, we take interpretability to mean the decodability of a latent skill from an agent's observed behavior---i.e., a policy is interpretable if it produces events and actions in a consistent or distinguishable manner.
While a flat policy is a black-box, since the action output is purely determined by the agent's observation input, the modularity of hierarchical models also provides an entry point for external control over the skills executed by AI teammates (e.g., execute the offense skill when it observes a human teammate doing so).

However, decomposing a global team objective such as ``scoring a goal'' into many sub-objectives for training a collection of skills is extremely difficult without expert knowledge, which may be hard to access for complex settings such as competitive team sports.
Manually crafting reward functions for each skill in high-dimensional state spaces involving numerous agents is also prone to misspecification and cause unintended behavior \citep{amodei2016concrete}.
Instead, we investigate a method for hierarchical agents in MARL to discover and learn a set of high-level latent skills. 
Agents should learn to cooperate by choosing effective combinations of skills with their teammates, and also dynamically choose skills in response to the state of the game.
In contrast to prior work in single-agent settings, where motion skills were discovered purely via an intrinsic reward \citep{achiam2018variational,eysenbach2018diversity}, MARL poses significant new challenges for skill discovery.
Merely discovering distinguishable individual motion in an open-ended multi-agent environment may be useless for a team objective.
While increasing the number of skills increases the chance that some are useful for a task \citep{eysenbach2018diversity}, doing so in the hierarchical multi-agent setting means exponentially increasing the size of a joint high-level action space and will exacerbate the difficulty of learning.

We present a method for training hierarchical policies with unsupervised skill discovery in cooperative MARL, with the following key technical and experimental contributions.
1) We construct a two-level hierarchical agent for MARL by defining a high-level action space as a set of latent variables.
Each agent consists of a high-level policy that chooses and sustains a latent variable for many time steps, and a low-level policy that uses both its observation and the selected latent variable to take primitive actions.
2) We use an extrinsic team reward to conduct centralized training of high-level policies for cooperation, while we use a combination of an intrinsic reward and the team reward to conduct decentralized training of low-level policies with independent reinforcement learning (RL).
This allows the use of powerful and general algorithms for cooperative MARL and single-agent RL to train high- and low-level policies, respectively.
3) We define the intrinsic reward as the performance of a decoder that predicts the ground truth latent variable from trajectories generated by low-level policies that were conditioned on the latent variables.
By dynamically weighting the intrinsic versus extrinsic reward, each low-level policy is trained to reach a balance between decodability and usefulness---it executes a skill, without the need for hand-designed skill-specific reward functions.
4) We applied this algorithm to a highly stochastic continuous state simulation of team sports and performed a detailed quantitative investigation of the learned behaviors.
Agents discover useful skills, that affect game events and determine low-level actions in distinct and interpretable ways, such as grouping together to steal possession from an opponent.
They learn to choose complementary skills among the team, such as when one agent camps near the opponent goal to get a rebound when its teammate makes a long-range shot attempt.
5) Our hierarchical agents perform higher than flat methods in ad-hoc cooperation when matched with teammates who follow policies that were not encountered in training.
This is an encouraging result for the possibility of human-AI cooperation.

\section{Related works}
\label{sec:related-works}

Building on the framework of options, temporally-extended actions, and hierarchical single-agent RL \citep{dayan1993feudal,sutton1999between,precup2000temporal,stolle2002learning},
early work on hierarchical MARL in discrete state spaces with hand-crafted subtasks \citep{makar2001hierarchical,ghavamzadeh2006hierarchical} showed that learning cooperation at the level of subtasks significantly speeds up learning over flat methods \citep{tan1993multi,littman1994markov,stone2000multiagent}.
Recent work built on deep reinforcement learning \citep{mnih2015human,silver2016mastering} to demonstrate hierarchical single-agent RL in high-dimensional continuous state spaces, using predefined subgoals \citep{kulkarni2016hierarchical}, end-to-end learning of options \citep{bacon2017option}, and latent directional subgoals \citep{vezhnevets2017feudal} in a two-level hierarchy.
In hierarchical MARL, different subtasks are chosen concurrently by all agents, whereas only a single subtask is chosen for each segment in single-agent hierarchical RL \citep{bacon2017option,vezhnevets2017feudal}.

Progress in hierarchical learning benefits from a complementary line of work on automatic subgoal discovery \citep{mcgovern2001automatic}.
Our work draws inspiration from variational option discovery \citep{achiam2018variational,gregor2016variational,eysenbach2018diversity}, which---in formal analogy with variational auto-encoders \citep{kingma2013auto}---trains a maximum-entropy policy encoder to map latent context vectors into trajectories from which the context can be recovered by a supervised decoder.
In contrast to prior work on single-agent skill discovery that focus on finding distinguishable behavior in simulated robotics environments, option discovery in cooperative MARL poses significant new demands:
1) individually distinguishable behaviors must be useful for the team objective;
2) hoping to discover useful skills by increasing the number of latent skills is impractical for the exponentially larger action space of MARL;
and 3) skills must be discovered in the actual multi-agent environment rather than in an isolated single-agent setting.

The key differences from recent work in hierarchical MARL \citep{tang2018hierarchical,ahilan2019feudal} are that we discover skills with an intrinsic reward instead of hand-crafting subtask-specific rewards \citep{tang2018hierarchical}, and our agents are on equal footing without a dedicated ``Manager'' \citep{ahilan2019feudal}.
A concurrent work on MARL with latent skills \citep{lee2020learning} require fully-centralized execution using global state information, while our method enables decentralized execution with local observations.
A complementary line of work learns \textit{role}-specific parameters and assignment of roles to agents with unique features, where each role is sustained for an entire episode \citep{wilson2010bayesian}, while our agents can dynamically choose skills multiple times in an episode.
We design our hierarchical agents using QMIX \citep{rashid2018a} and independent DQN \citep{tan1993multi,mnih2015human}; other decentralized cooperative MARL \citep{hernandez2018multiagent} and single-agent RL \citep{sutton2018reinforcement} algorithms are equally applicable.



\section{Methods}
\label{sec:methods}

We present a method for fully-cooperative hierarchical MARL, whereby independently-acting agents learn to cooperate using latent skills that emerge from a combination of intrinsic and extrinsic rewards.
Inspired by training practices of real world professional sports teams, we create our method within the paradigm of centralized training with decentralized execution \citep{bernstein2002complexity}.
For ease of exposition and intuition, we assume all agents have the same observation space and action space; nevertheless they take individual actions based on  individual observations.
In the rest of this section, we define the objective of hierarchical MARL with skill discovery, describe our method to solve the optimization problem, and discuss practical implementation techniques for effective learning.

\begin{figure*}[t]
    \centering
    \includegraphics[width=0.65\linewidth]{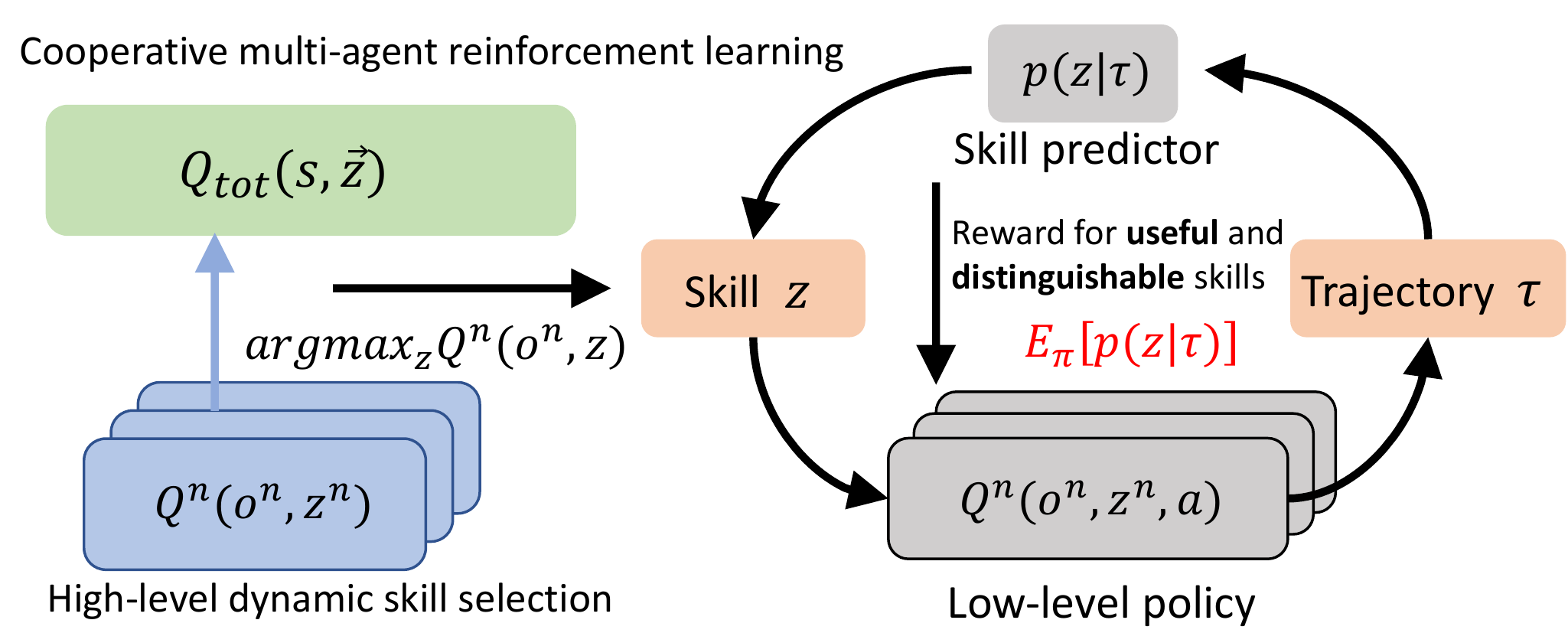}
    \caption{Hierarchical MARL with unsupervised skill discovery. At the high level (left), the extrinsic team reward is used to train a centralized action-value function $Q_{\text{tot}}(s,\zbf)$ that decomposes into individual utility functions $Q^n(o^n,z^n)$ for decentralized selection of latent skill variables $z$. At the low level (right), skill-conditioned action-value functions $Q^n(o^n,z^n,a^n)$ take primitive actions independently. Trajectories $\tau$ generated under each $z$ are collected into a dataset $\Dcal = \lbrace (z, \tau) \rbrace$, which is used to train a skill decoder $p(z|\tau)$ to predict $z$ from $\tau$. The probability of selected skills under $p(z|\tau)$ is the intrinsic reward for low-level $Q^n$.}
    \label{fig:architecture}
\end{figure*}

\subsection{Combining centralized and decentralized training in hierarchical MARL}

We describe a two-level hierarchical MARL setup for training $N$ agents, labeled by $n \in [N]$, as follows.
Let $\Zcal$ denote a set of latent variables $z$, each of which corresponds to a \textit{skill}.
In this work, we use a finite set of latent variables with one-hot encoding; it is possible to generalize $\Zcal$ to be a learned continuous embedding space \citep{achiam2018variational}.
We treat $\Zcal$ as the \textit{action space} for high-level policies\footnote{Without loss of generality, and for consistency with our algorithm implementation below, we use the notation for deterministic policies in this paper.} $\mu^n \colon \Ocal \mapsto \Zcal, \forall n \in [N]$, each of which maps from an agent's observation $o^n \in \Ocal$ to a choice of skill $z^n \in \Zcal$.
Each choice of $z^n$ is sustained for $t_{\text{seg}}$ time steps: letting $T = K t_{\text{seg}}$
denote the length of an episode, there are $K$ time points at which a high-level skill selection is made (see \Cref{subsec:traj-processing}).
Conditioned on a chosen latent skill and given an agent's observation, a low-level policy $\pi^n \colon \Ocal \times \Zcal \mapsto \Acal$ outputs a primitive action $a^n$ in a low-level action space $\Acal$.
Each $z \in \Zcal$ and the latent-conditioned policy $\pi^n(\cdot;z^n)$ is a skill, in accord with terminology in the literature \citep{gregor2016variational,eysenbach2018diversity,achiam2018variational}.
Let boldface $\mubf, \pibf$, and $\abf$ denote the joint high-level policy, joint low-level policy, and joint action, respectively.
Let $(\cdot)^{-n}$ denote a joint quantity for all agents except agent $n$.
At the high level, $\mubf$ learns to select skills to optimize an extrinsic team reward function $R \colon \Scal \times \lbrace \Acal \rbrace_{n=1}^N \mapsto \Rbb$ that maps global state and joint action to a scalar reward.
At the low level, $\lbrace \pi^n \rbrace_{n=1}^N$ learn to choose primitive actions to produce useful and decodable behavior by optimizing a low-level reward function $R_L$.
Combining the learning at both levels, we view hierarchical MARL as a bilevel optimization problem \citep{colson2007overview}: 
\begin{align}
    &\max_{\mubf, \pibf} \Ebb_{\zbf \sim \mubf,P} \left[ \Ebb_{s_t,a_t \sim \pibf,P} \left[ \sum_{t=1}^T \gamma^t R(s_t,\abf_t) \right] \right] \label{eq:objective-high} \\ 
    &\pi^n \in \argmax_{\pi} \Ebb_{
    \zbf \sim \mubf, P}
    \left[ \sum_{k=1}^K \Ebb_{\tau^n_k \sim 
    \pibf, P}
    \left[ 
    R_L(z^n_k,\tau^n_k)
    \right] \right], \forall n \in [N] \label{eq:objective-low}
\end{align}
where $\tau^n_k$ is the $k$-th trajectory segment that consists of a sequence of observations by agent $n$, $P$ denotes the environment transition probability $P(s_{t+1}|s_t,\abf)$,
and 
$R_L(z^n, \tau^n) := \sum_{(s_t,a_t) \in \tau^n} R_L(z^n, s_t, a_t)$ 
denotes the sum of agent $n$'s low-level rewards along trajectory $\tau^n$.
This may also be viewed as a general-sum meta-game between a $\mubf$-player and another $\pibf$-player.
When $R_L$ is the extrinsic team reward, we have a fully-cooperative meta-game, while the other extreme is where $R_L$ solely promotes decodability. 
Our approach, explained in \Cref{subsec:skill-discovery}, lies in between these extremes to strike a balance between usefulness and decodability. 

It is difficult to solve \eqref{eq:objective-high}-\eqref{eq:objective-low} exactly in high-dimensional continuous state spaces.
Furthermore, we adjust $R_L$ dynamically to promote skill predictability (see \Cref{subsec:skill-discovery}).
Instead, we approach it using powerful algorithms for MARL and RL.
First, we use centralized MARL algorithms to train high-level policies $\mubf$ for cooperative high-level skill selection.
While cooperative behavior may emerge from flat policies trained by a team reward \citep{liu2019emergent}, explicitly training high-level skill-selection policies allows external control over the choice of skills performed (by fixing a latent variable), and subsequent analysis of the behavior for each skill.
Second, we apply independent RL to train low-level policies $\lbrace \pi^n \rbrace_{n=1}^N$, each conditioned on a skill selected by the agent's corresponding high-level policy, to take primitive actions to optimize $R_L(z^n, \tau^n)$ (defined below in \Cref{subsec:skill-discovery}).
This reflects the fact that human players in team sports can master skills individually outside of team practice.

\subsection{Skill discovery via dynamically weighted decoder-based intrinsic rewards}
\label{subsec:skill-discovery}

We define the low-level reward by first introducing a skill decoder $p_{\psi}(z^n|\tau^n)$ that predicts the ground truth latent skill $z^n$ that was used in the low-level policy $\pi(\cdot;z^n)$ that generated the trajectory $\tau^n$.
The decoder is trained using a dataset $\Dcal = \lbrace (z, \tau) \rbrace$ of skill-trajectory pairs, where each consists of the $z$ chosen by a high level policy and the corresponding trajectory $\tau$ generated by the low level policy given $z$, over all agents.
$\Dcal$ is accumulated in an online manner during training.
Hence, training $p_{\psi}$ alone can be viewed as a supervised learning problem where we have access to the ground truth ``label'' $z$ associated with each ``datapoint'' $\tau$.

We define the intrinsic reward $R_I(z^n_k, \tau^n_k)$ for agent $n$'s $k$-th trajectory segment $\tau^n_k$ via the prediction performance of the skill decoder on the tuple $(z^n_k, \tau^n_k)$.
Agent $n$ receives this scalar reward upon generating the segment $\tau^n_k$.
The key intuition is that a skill in many complex fully-cooperative team games can be inferred from the trajectory of primitive actions that implement the skill \citep{le2017coordinated,hernandez2018multiagent}.
For example, any agent who executes a defensive subtask in soccer will move toward opponents in a consistent way that mainly depends on its own observations, with only weak dependence on the behavior of other physically distant agents\footnote{As a first step, we do not include higher-order skills that involve coordinated behavior of two or more agents. Our method can be extended to higher-order skills by associating multiple agents' concurrent trajectories with a single skill.}.
This intrinsic reward encourages the generation of distinguishable behavior for different skills, since only by doing so can the low-level policy produce sufficiently distinct ``classes'' in the dataset $\Dcal$ for the decoder to achieve high prediction performance. 
Hence we define the low-level reward $R_L$ as a combination of team reward $R$ and intrinsic reward $R_I$:
\begin{align}
    R_L(z^n,\tau^n) &:= \alpha \sum_{s_t,\abf_t \in \tau^n} \gamma^t R(s_t,\abf_t) + (1-\alpha) R_I(z^n, \tau^n) \label{eq:reward-lower}\\ 
    \text{where } R_I &:= p_{\psi}(z^n|\tau^n) \label{eq:reward-intrinsic}
\end{align}
$\alpha \in \Rbb$ is a \textit{dynamic} weight (specified below) that determines the amount of intrinsic versus environment reward.
In contrast to prior work on single-agent option discovery that do not use an extrinsic reward \citep{gregor2016variational,eysenbach2018diversity,achiam2018variational}, 
we take advantage of the team reward in MARL to guarantee that skills are useful for team performance, and rely on the intrinsic reward only to promote the association of latent variables with predictable behavior.
This ensures that low-level policies, when conditioned on different latent variables, produce trajectories that are 1) sufficiently different to allow decoding of the latent variable, and 2) useful for attaining the true game reward---e.g. ``attack opponent net'' and ``defend own net''.
We decrease $\alpha$ from 1.0 to $\alpha_{\text{end}}$ via an automatic curriculum in which $\alpha$ decreases by $\alpha_{\text{step}}$ only when the performance (e.g., win rate) in evaluation episodes, conducted periodically during training, exceeds a threshold $\alpha_{\text{threshold}}$.
At high $\alpha$, low-level policies learn independently to maximize the team reward by taking useful actions, some of which can be composed into interpretable behavior.
As $\alpha$ decreases and the skill decoder associates trajectories with latent variables, the low-level policy is increasingly rewarded for generating easily decodable modes of behavior when conditioned on different $z$. 
A high $\alpha_{\text{threshold}}$ can be more suitable for highly stochastic games (see \Cref{subsec:performance}), so that the weight on the intrinsic reward increases later during training, after agents have learned to take useful actions.

\begin{algorithm}[t]
\caption{Hierarchical MARL with unsupervised skill discovery}
\label{alg:sd}
\begin{algorithmic}[1]
\Procedure{Algorithm}{}
\State Initialize high-level $Q_{\phi}$, low-level $Q_{\theta}$, decoder $p_{\psi}$, high-level replay buffer $\Bcal_H$, low-level replay buffer $\Bcal_L$, and trajectory-skill dataset $\Dcal$
\For{each episode}
    \State $s_t, \obf_t = $ env.reset()
    \State Initialize trajectory storage $\lbrace \tau^n \rbrace_{n=1}^N$ of max length $t_{\text{seg}}$
    \For{each step $t = 1,\dotsc,T$ in episode}
        \If{$t \mod t_{\text{seg}} = 0$}
            \If{$t > 1$}
                \State Compute $\tilde{R}_t := \gamma^{t_{\text{seg}}} * \sum_{k=0}^{t_{\text{seg}}} R_{t-k}$
                \State Store $(s_{t-t_{\text{seg}}}, \obf_{t-t_{\text{seg}}}, \zbf, \tilde{R}_t, s_t, \obf_t)$ into $\Bcal_H$
                \For{each agent $n$}
                    \State Store $(z^n, \tau^n)$ into $\Dcal$
                    \State Compute intrinsic reward $R^n_I$ using \eqref{eq:reward-intrinsic}
                \EndFor
            \EndIf
            \State Select new $z^n$ by $\epsilon$-greedy$(Q^n_{\phi}(o^n,z)), \forall n \in [N]$
            \If{$\text{\# (high level steps)} \mod t_{\text{train}} = 0$}
                \State Update $Q_{\phi}(s,\zbf)$ using $\Bcal_H$ and \eqref{eq:loss-q-high} 
            \EndIf
        \EndIf
        \State Get $a^n_t$ from $\epsilon$-greedy$(Q(o^n_t,z^n_t,a))$ for each agent
        \State $s_{t+1}, \obf_{t+1}, R_t$ = env.step($\abf_t$)
        \State Compute $R_L^n := \alpha R_t + (1-\alpha) R^n_I$ for each agent
        \State For all agents, store $(o^n_t, a^n_t, R_L^n, o^n_{t+1}, z^n)$ into low-level replay buffer $\Bcal_L$, and append $o^n_t$ to trajectory $\tau^n$
        \If{\# (low-level steps) $\mod t_{\text{train}} = 0$}
            \State Update $Q_{\theta}(o^n,z^n,a^n)$ using $\Bcal_L$ and \eqref{eq:loss-q-low}
        \EndIf
    \EndFor
    \If{size of $\Dcal \geq N_{\text{batch}}$}
        \State Update decoder $p_{\psi}(z|\tau)$ using $\Dcal$, then empty $\Dcal$
    \EndIf
    \If{evaluation win rate exceeds $\alpha_{\text{threshold}}$}
        \State $\alpha \leftarrow \max(\alpha_{\text{end}}, \alpha - \alpha_{\text{step}})$
    \EndIf
\EndFor
\EndProcedure
\end{algorithmic}
\end{algorithm}

\subsection{Algorithm}

\Cref{alg:sd} is our approach to the optimization problem \cref{eq:objective-high,eq:objective-low}, with skill discovery based on \cref{eq:reward-lower}.
We initialize replay buffers $\Bcal_H, \Bcal_L$ for both levels of the hierarchy, for off-policy updates in similar style to DQN \citep{mnih2015human}, and initialize a dataset $\Dcal$ for the decoder (line 2).
At the $k$-th high-level step, which occurs once for every $t_{\text{seg}}$ primitive time steps (line 6), we compute the SMDP reward $\tilde{R}_t := \sum_{i=0}^{t_\text{seg}-1} \gamma^i R(s_{t-i},\abf_{t-i})$ for the high-level policy (line 8) \citep{sutton1999between}. 
Each agent computes its reward and independently selects a new skill to execute for the next high-level step (lines 12-13).
We periodically take gradient steps to optimize the high level cooperative skill-selection objective \eqref{eq:objective-high} (lines 15-17), using QMIX \citep{rashid2018a} to train a centralized Q-function $Q^{\text{tot}}_{\phi}(s_t,\zbf)$ via minimizing the loss:
\begin{align}
    \Lcal(\phi) &:= \Ebb_{\mubf, \pibf} \left[ \frac{1}{2} \left( y_k - Q^{\text{tot}}_{\phi}(s_k, \zbf_k) \right)^2 \right] \label{eq:loss-q-high} \\
    y_k &:= \tilde{R}_k + \gamma Q^{\text{tot}}_{\phi}(s_{k+1}, \zbf')|_{\lbrace z'^n = \argmax_{z^n} Q^n_{\phi}(o^n_{k+1}, z^n) \rbrace_{n=1}^N} \label{eq:td-target-high}
\end{align}
$Q^{\text{tot}}_{\phi}$ is a non-linear function (e.g., neural network) that is monotonic in individual utility functions $Q^n_{\phi}, n \in [N]$, and we denote $\mubf$ as the collection of greedy policies induced by $Q^n_{\phi}$.
The hypernetwork of QMIX enforces $\partial Q^{\text{tot}}_{\phi}/\partial Q^n_{\phi} > 0$, which is a sufficient condition for a global $\argmax$ to be achieved via decentralized $\argmax$, i.e., $\argmax_{\zbf} Q^{\text{tot}}_{\phi}(\cdot, \zbf) = \lbrace \argmax_{z^n} Q^n_{\phi}(\cdot, z^n) \rbrace_{n=1}^N$.
This allows centralized training with decentralized skill selection.
In general, one can choose from a diverse set of cooperative MARL algorithms with decentralized execution \citep{foerster2018counterfactual,sunehag2018value,son2019qtran,yang2018cm3}.

Conditioned on the choices of skills, each agent independently executes primitive actions at every low-level time step (lines 19-20), using the greedy policy $\pi^n$ induced by low-level Q-functions $Q^n_{\theta}(o^n_t,z^n_t,a^n)$.
We periodically take gradient steps to optimize the low level objective \eqref{eq:objective-low} (lines 23-25), by using independent DQN \citep{tan1993multi,watkins1992q,mnih2015human} to optimize $Q^n_{\theta}$ via minimizing the loss:
\begin{align}
    \Lcal(\theta) &:= \Ebb_{\mubf,\pibf} \left[ \frac{1}{2}\left( y^n_t - Q^n_{\theta}(o^n_t,z^n,a^n_t) \right)^2 \right] \label{eq:loss-q-low} \\
    y^n_t &:= R_L(z^n, \tau^n) + \gamma \max_{a^n} \hat{Q}^n_{\theta}(o^n_{t+1},z^n,a^n), \forall n \in [N] \label{eq:td-target-low}
\end{align}
$\pibf$ denotes the collection of greedy policies induced by all $Q^n_{\theta}$.
The low level reward $R_L$ includes the contribution of the intrinsic reward $R_I$ only at the final time step of each length-$t_{\text{seg}}$ trajectory segment, i.e., at every high-level step.
$\hat{Q}$ is a target network \citep{mnih2015human}.

Once $N_{\text{batch}}$ number of $(z^n, \tau^n)$ are collected into the dataset $\Dcal$ (lines 11, 27-29), the skill decoder $p_{\psi}(z|\tau)$ is trained to predict $z$ given $\tau$ via supervised learning on $\Dcal$ by minimizing a standard cross-entropy loss.
Each chosen $z^n$ acts as the class label for the corresponding trajectory $\tau^n$.
Periodically, we evaluate the agents' performance (e.g., win rate) in seperate evaluation episodes; if performance exceeds $\alpha_{\text{threshold}}$, we decrease the weight $\alpha$ by $\alpha_{\text{step}}$ with lower bound $\alpha_{\text{end}}$ (\Cref{subsec:skill-discovery}).
While it is extremely challenging to provide theoretical guarantees for hierarchical methods, especially due to the need for nonlinear function approximation to tackle high-dimensional continuous state spaces, simultaneous optimization in hierarchical RL has shown promising practical results \citep{bacon2017option,vezhnevets2017feudal}.



\subsection{Trajectory segmentation and compression}
\label{subsec:traj-processing}

Hierarchical MARL requires agents to change their choice of skills dynamically at multiple times within an episode, such as in response to a change of ball possession in soccer.
This means we use partial segments instead of full episode trajectories for skill discovery, in contrast to the single-agent case \citep{gregor2016variational,eysenbach2018diversity,achiam2018variational}. 
At first glance, using a fixed time discretization hyperparameter $t_{\text{seg}}$ for segmentation may pose difficulties for the skill decoder, such as when a segment contains qualitatively different behavior that should correspond to different skills.
We address this issue by using the time points at which the high-level policy chooses a new set of skill assignments as the segmentation.
Hence, $\pibf$ learns to generate trajectory segments in between the time points, and $p_{\psi}$ learns to associate these segments with the chosen latent variables.
We synchronize the time points of all agents' high-level skill choice, and all skills are sustained for $t_\text{seg}$ low-level steps.
This corresponds to a special case of the ``any'' termination scheme, which is dominant over other termination schemes considered in \citep{rohanimanesh2003learning}.
A practical approach is to define a range of values based on domain knowledge (e.g., average duration of a player's ball possession) and include it in hyperparameter search.
Agents can still learn skills that require more than $t_\text{seg}$ steps, by sustaining the same skill for multiple high-level steps.

Building on \citep{achiam2018variational}, we preprocess each trajectory before using it as input to the decoder.
We downsample by retaining every $k_{\text{skip}}$ steps, which filters out low-level noise in stochastic environments.
We use the element-wise difference between the downsampled observation vectors.
This discourages the possibility that more than one skill exhibits stationary behavior (e.g., camping at different regions of a field), as the difference will be indistinguishable for the decoder and result in low intrinsic reward.
We reduce the dimension of observation vectors for the decoder by removing entries corresponding to all other agents, while retaining game-specific information (e.g., ball possession).
Hence an agent's own trajectory must contain enough information for decoding the latent skill variable.




\section{Experimental Setup}
\label{sec:setup}

Our experiments demonstrate that the proposed method discovers interpretable skills that are useful for high-level strategies and has potential for human-AI cooperation in team sports games\footnote{Code for experiments is available at \url{https://github.com/011235813/hierarchical-marl}}.
We contribute evidence that hierarchical MARL with unsupervised skill discovery can meet or exceed the performance of non-hierarchical methods in high-dimensional environments with only a global team reward.
We describe the simulation setup in \Cref{subsec:sts2}
and provide full implementation details of all methods in \Cref{subsec:implementation}.

\subsection{Simple Team Sports Simulator}
\label{subsec:sts2}

The Simple Team Sports Simulator (STS2) captures the high-level rules and physical dynamics of general $N$ versus $N$ team sports while abstracting away fine-grained details that do not significantly impact strategic team play \citep{sts2_ea_2020,zhao2019multi}.
Stochasticity of ball possession and goals makes STS2 a challenging environment for MARL.
Complementary to 3D simulations such as \citet{kurach2019google} that require massively parallelized training, STS2 is a lightweight benchmark where MARL agents can outperform the scripted opponent team within hours on a single CPU.
We train in 3v3 mode against the scripted opponent team for 50k episodes. 
Each episode terminates either upon a goal or a tie at 500 time steps.

\textbf{State.}
We define a state representation that is invariant under 180 degree rotation of the playing field and switch of team perspective.
For one team, the state vector has the following components, making up total dimension 34:
normalized position of the player with possession relative to the goal, and its velocity;
a 1-hot vector indicating which team or opponent player has possession;
for each team and opponent player, its normalized position and velocity.

\textbf{Observation.}
Each agent has its own egocentric observation vector with the following components, making up total dimension 31:
normalized position and velocity of the player with possession relative to this agent;
a binary indicator of whether this agent has possession;
a binary indicator of whether its team has possession;
its normalized position and its velocity;
relative normalized position of each teammate, and their relative velocities;
a binary indicator of whether the opponent team has possession;
relative normalized position of each opponent player, and their relative velocities.

\textbf{Action.}
The low-level discrete set of actions consists of: do-nothing, shoot, pass-1, ... , pass-N, down, up, right, left. Movement and shoot directions are relative to the team's field side.
If the agent does not have possession and attempts to shoot or pass, or if it has possession and passes to itself, it is forced to do nothing. 

\textbf{Reward.}
The team receives reward $+1$ for scoring, $-1$ when the opponent scores, $\pm 0.1$ on the single step when it regains possession from, or loses possession to, the opponent.
We include a reward of $\pm 1/(2 * \text{max steps per episode})$ for having or not having possession.

\textbf{Game events.}
We define a set of game events, which are frequently used for analyzing team sports \citep{franks2015characterizing}, to quantify the effect of skills.
Goals: agent scored a goal, upon which an episode ends.
Offensive rebound: agent's team made a shot attempt, which missed, and the agent retrieved possession.
Shot attempts: agent attempted to score a goal.
Made or received pass: agent made (received) a successful pass to (from) a teammate.
Steals: agent retrieved possession from an opponent by direct physical contact.

\subsection{Implementation and baselines}
\label{subsec:implementation}

We use parameter-sharing among all agents, as is standard for homogeneous agents in cooperative MARL \citep{hernandez2018multiagent}.
For function approximation, we use fully-connected neural networks without recurrent units since the game is fully observable.
Each component is depicted in \Cref{fig:architecture}.
The low-level Q-function has two hidden layers, each with 64 units, and one output node per action.
The high-level Q-function is a QMIX architecture: the individual utility function has two layers with 128 units per layer, and one output node per skill.
Utility values of all agents are passed into a mixer network, whose non-negative weights in two hidden layers are generated by hypernetworks of output dimension 64, and whose final output is a single global Q value (see \citep{rashid2018a}).
The skill decoder is a bidirectional LSTM \citep{schuster1997bidirectional} with 128 hidden units in both forward and backward cells, whose outputs are mean-pooled over time and passed through a softmax output layer to produce probabilities over skills.
We use batch size $N_{\text{batch}}=1000$ to train the decoder;
$\epsilon$-greedy exploration at both high and low levels with $\epsilon$ decaying linearly from 0.5 to 0.05 in 1e3 episodes;
replay buffers $\Bcal_H$ and $\Bcal_L$ of size 1e5; learning rate 1e-4; and discount $\gamma= 0.99$.
High and low level action-value functions are trained using minibatches of 256 transitions every 10 steps at the high and low levels, respectively.
Target networks \citep{mnih2015human} are updated after each training step with update factor 0.01.
We conduct 20 episodes of evaluation once every 100 training episodes.
We experimented with 4 and 8 latent skills, $t_{\text{seg}} = 10$, and let $\alpha$ decay from 1.0 to a minimum of 0.6 by $\alpha_{\text{step}} = 0.01$ whenever average win rate during evaluation exceeds $\alpha_{\text{threshold}} = 70\%$.
We process trajectory segments as described in \Cref{subsec:traj-processing} with $k_{\text{skip}}= 2$.

As we instantiate our general method using QMIX \citep{rashid2018a} at the high level and independent Q-learning (IQL) \citep{tan1993multi,mnih2015human} at the low level, we compare performance with these two baselines to demonstrate that the new hierarchical architecture maintains performance while gaining interpretability.
QMIX uses the same neural architecture as our method, except that the individual utility function outputs action-values for primitive actions instead of action values for high-level skills.
IQL uses a two-layer Q-network with 128 units per layer.
We first performed a coarse manual search for hyperparameters of QMIX and IQL, and used the same same values for the corresponding subset of hyperparameters in our method.
Additional hyperparameters ($\alpha_{\text{threshold}}, \alpha_{\text{step}}$, and $t_{\text{seg}}$) in our method were chosen from a coarse manual search, and we show results on hyperparameter sensitivity.
We also compared with a variant of our method that uses two hand-scripted subtask reward functions with the same hierarchical architecture.
An agent with subtask 1 gets reward +1 for making a goal when having possession; an agent with subtask 2 gets +1 for stealing possession from an opponent.
These \textit{individual} rewards mitigate the difficult problem of multi-agent credit assignment, and so this variant gives a rough indication of maximum possible win rate against the scripted opponent team.

\begin{figure*}[t]
\centering
\begin{subfigure}[t]{.19\linewidth}
    \centering
    \includegraphics[width=\linewidth]{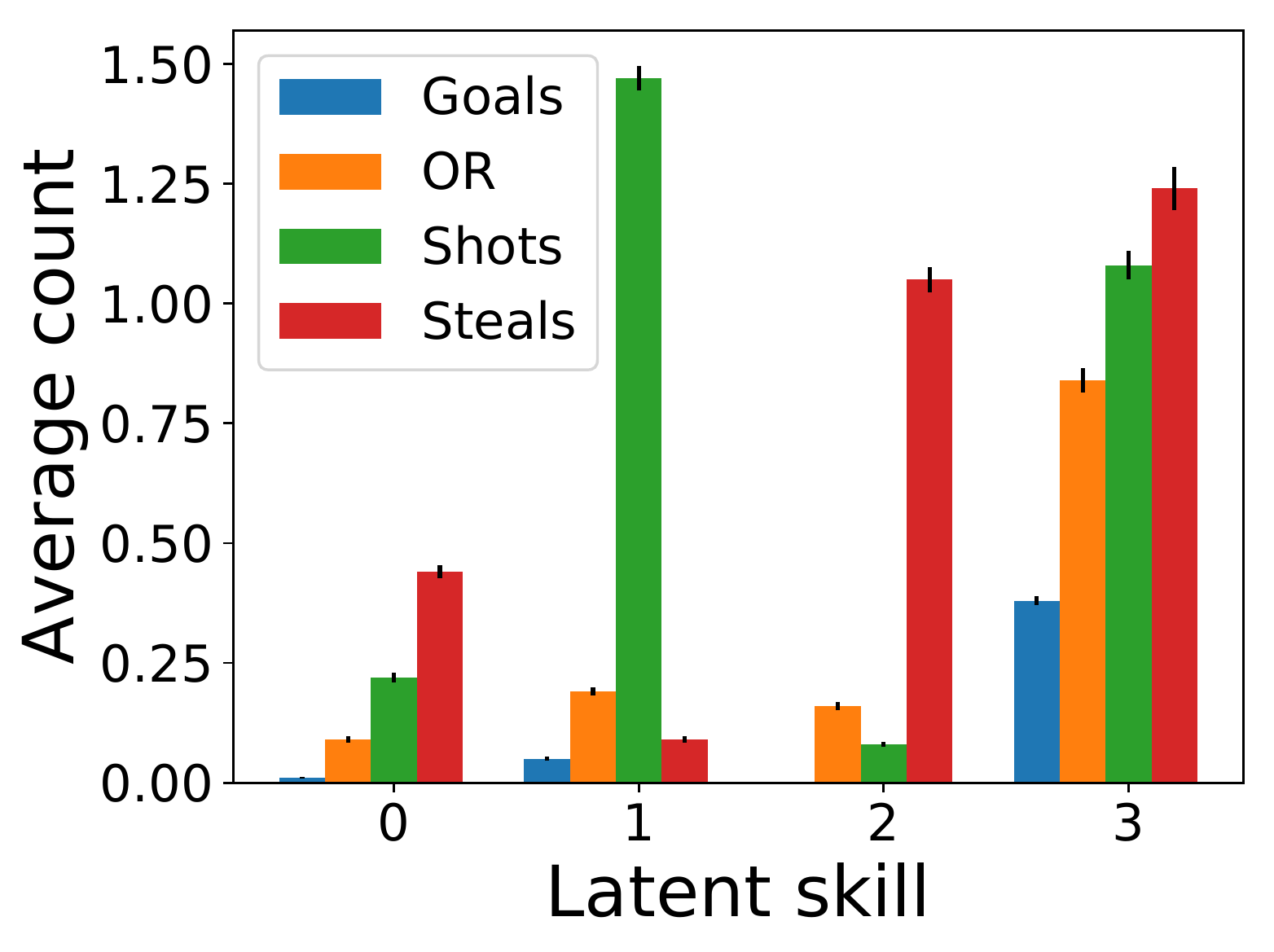}
    \caption{Event distribution}
    \label{fig:events-4}
\end{subfigure}
\hfill
\begin{subfigure}[t]{0.19\linewidth}
  \centering
  \includegraphics[width=0.98\linewidth]{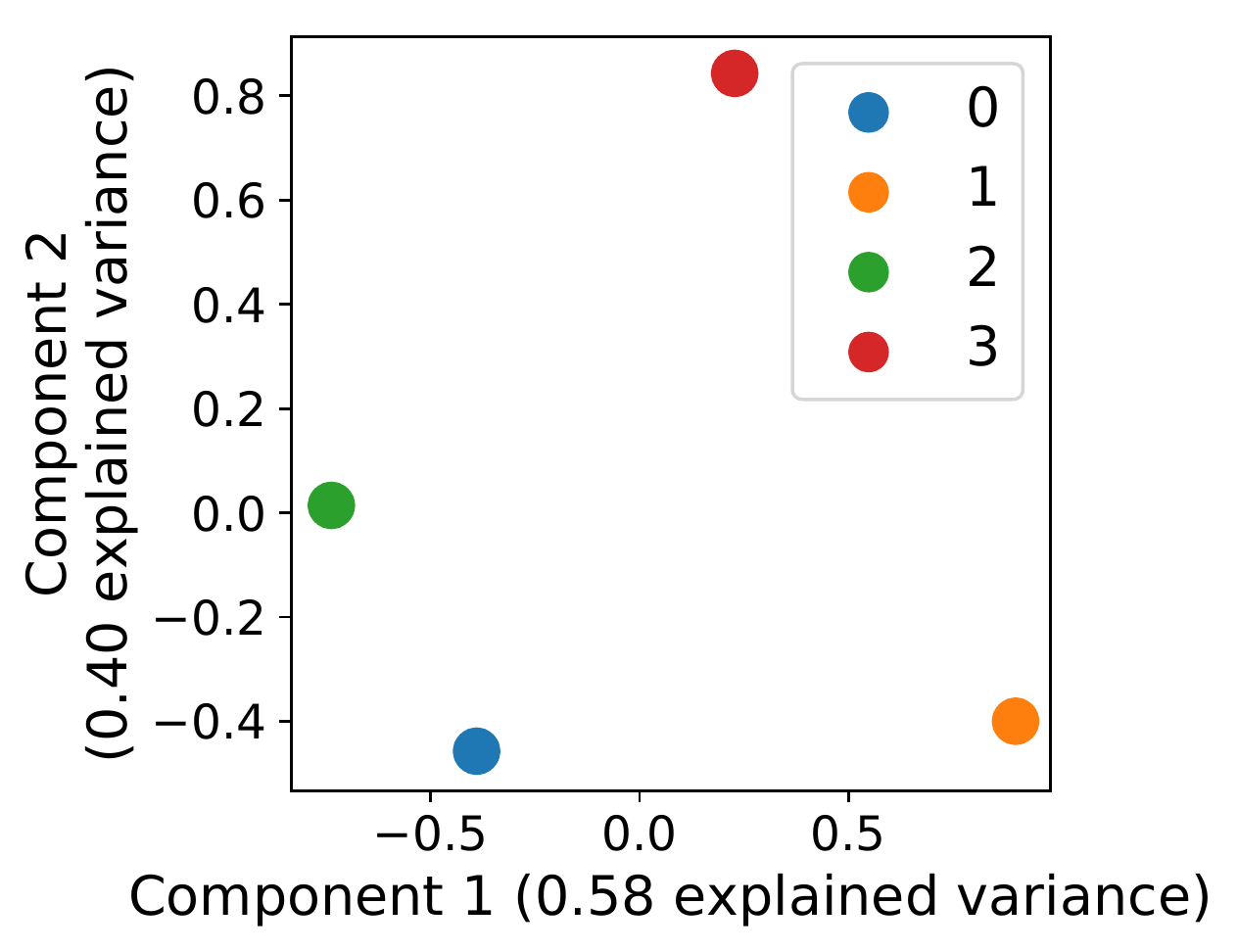}
  \caption{PCA of events}
  \label{fig:PCA-events-4}
\end{subfigure}
\hfill
\begin{subfigure}[t]{.19\linewidth}
  \centering
  \includegraphics[width=\linewidth]{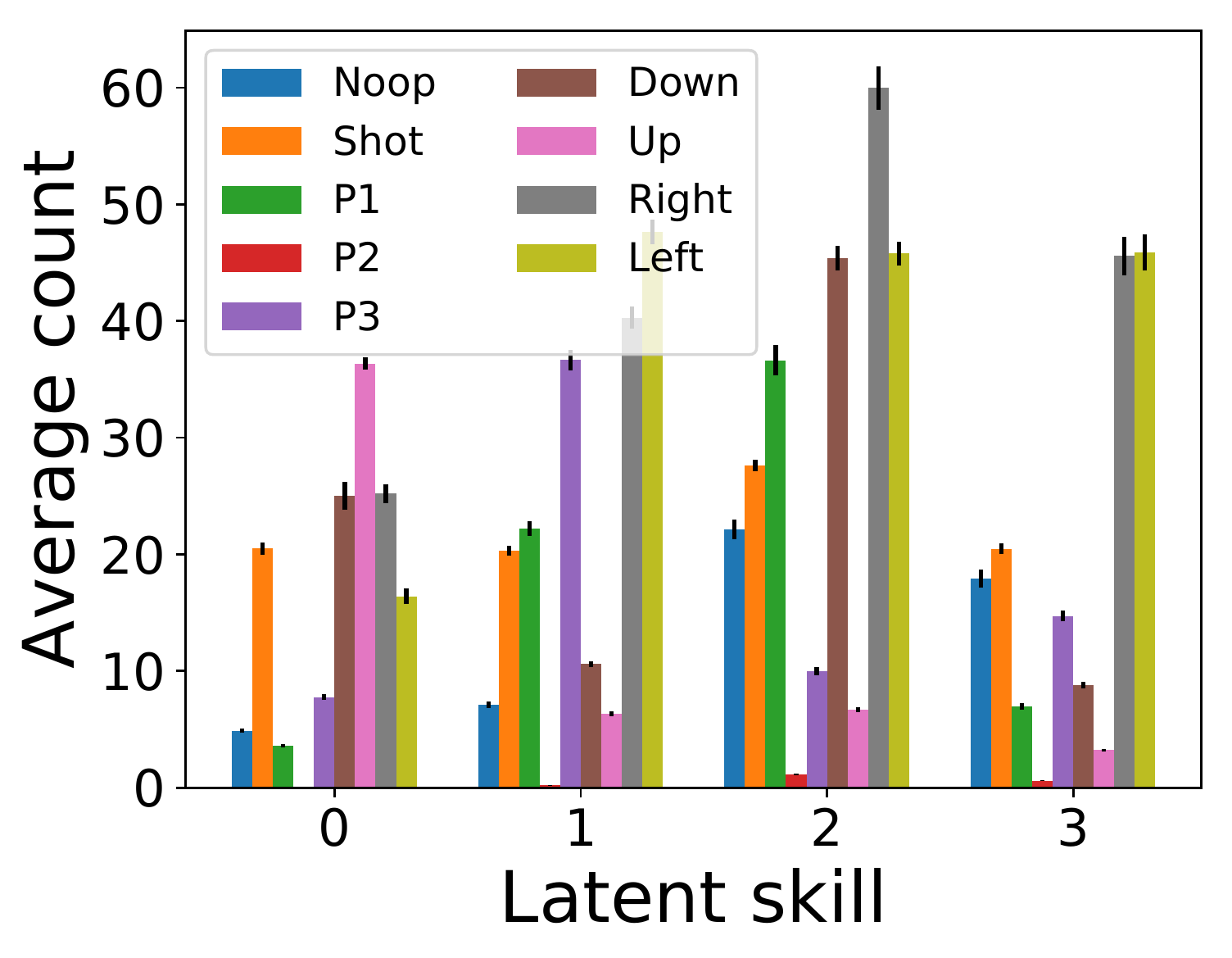}
  \caption{Action distribution}
  \label{fig:actions-4}
\end{subfigure}
\hfill
\begin{subfigure}[t]{0.19\linewidth}
  \centering
  \includegraphics[width=0.98\linewidth]{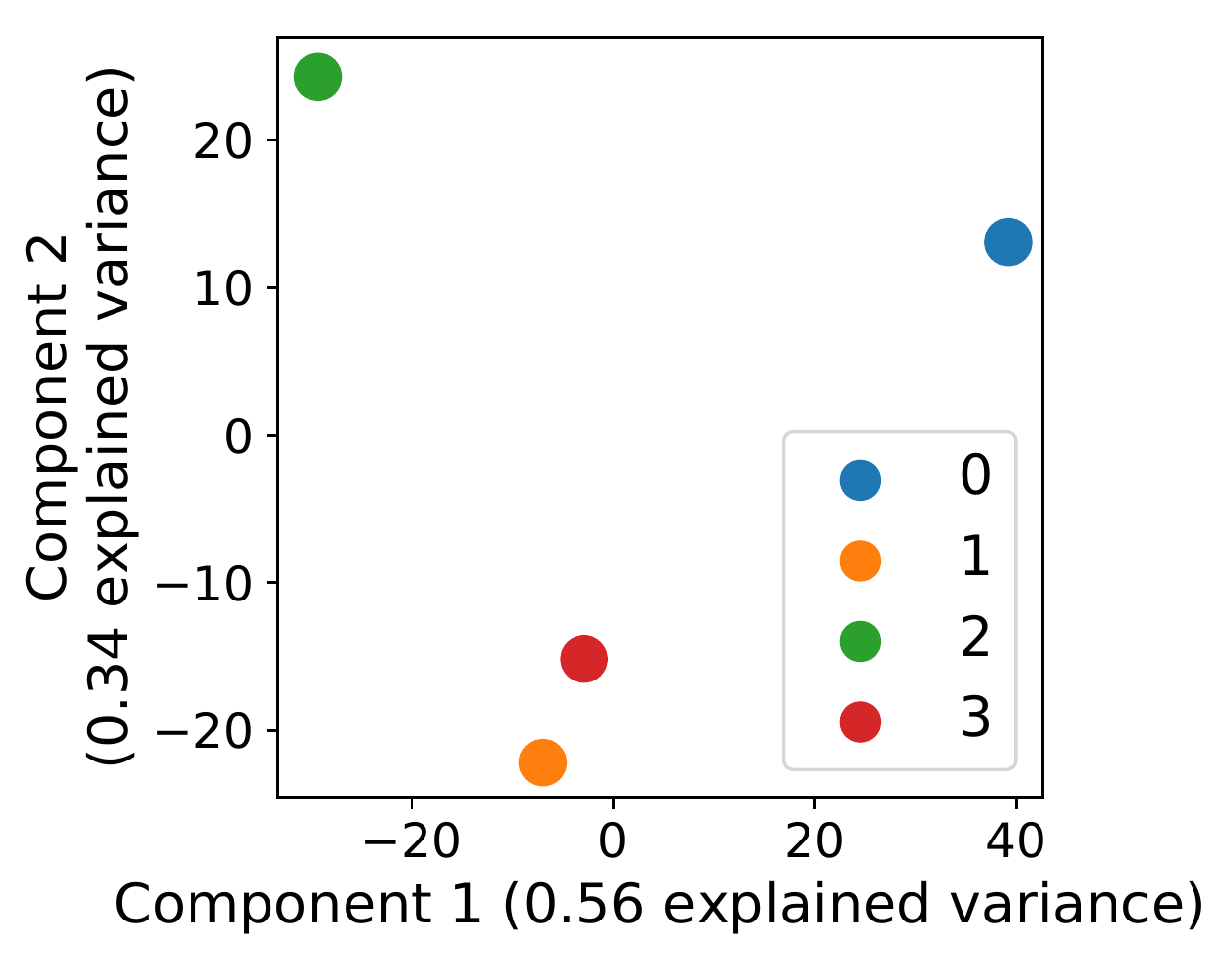}
  \caption{PCA of actions}
  \label{fig:PCA-actions-4}
\end{subfigure}
\hfill
\begin{subfigure}[t]{0.19\linewidth}
  \centering
  \includegraphics[width=\linewidth]{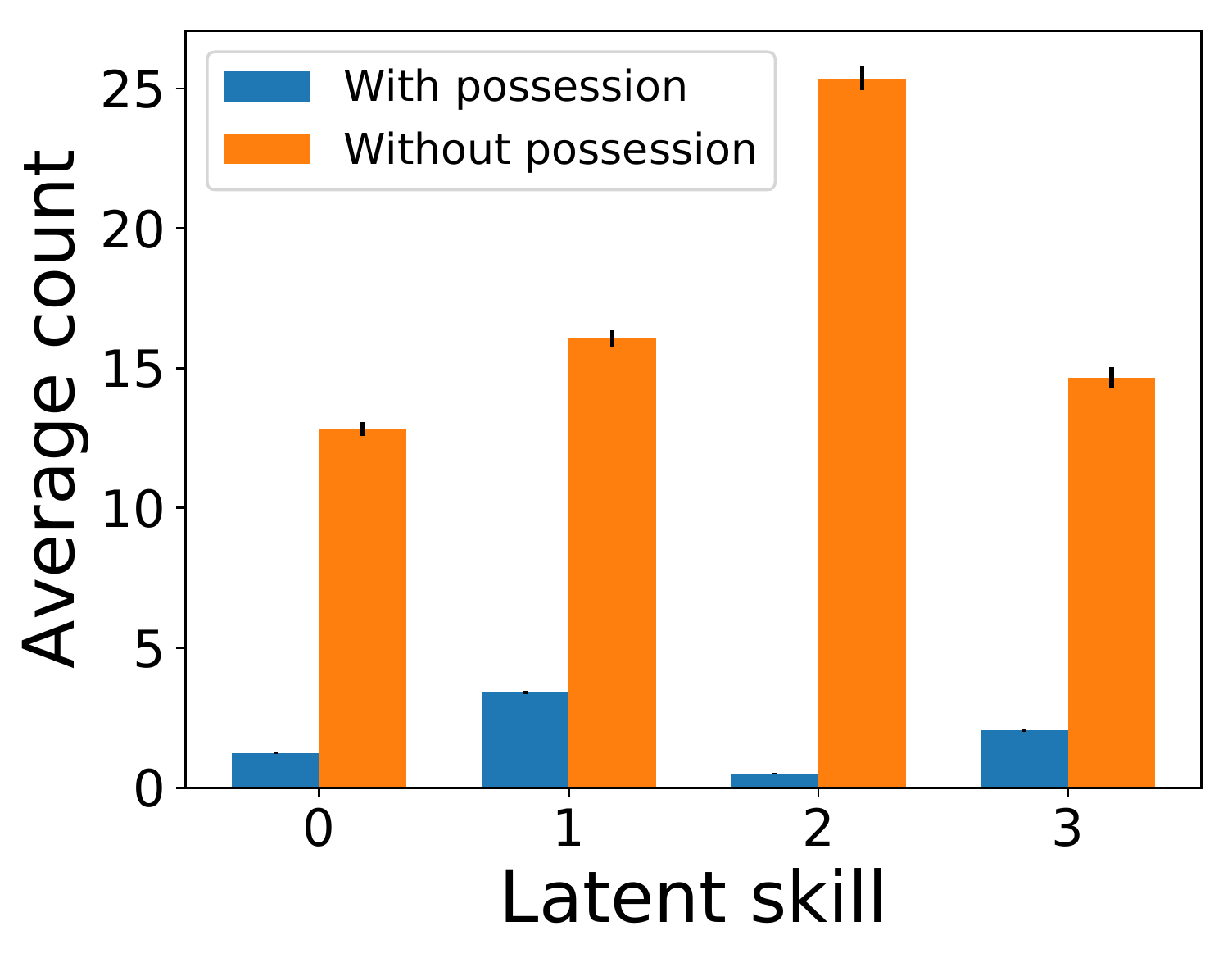}
  \caption{Overall skill usage}
  \label{fig:skills-4}
\end{subfigure}

\begin{subfigure}[t]{0.34\linewidth}
    \centering
    \includegraphics[width=\linewidth]{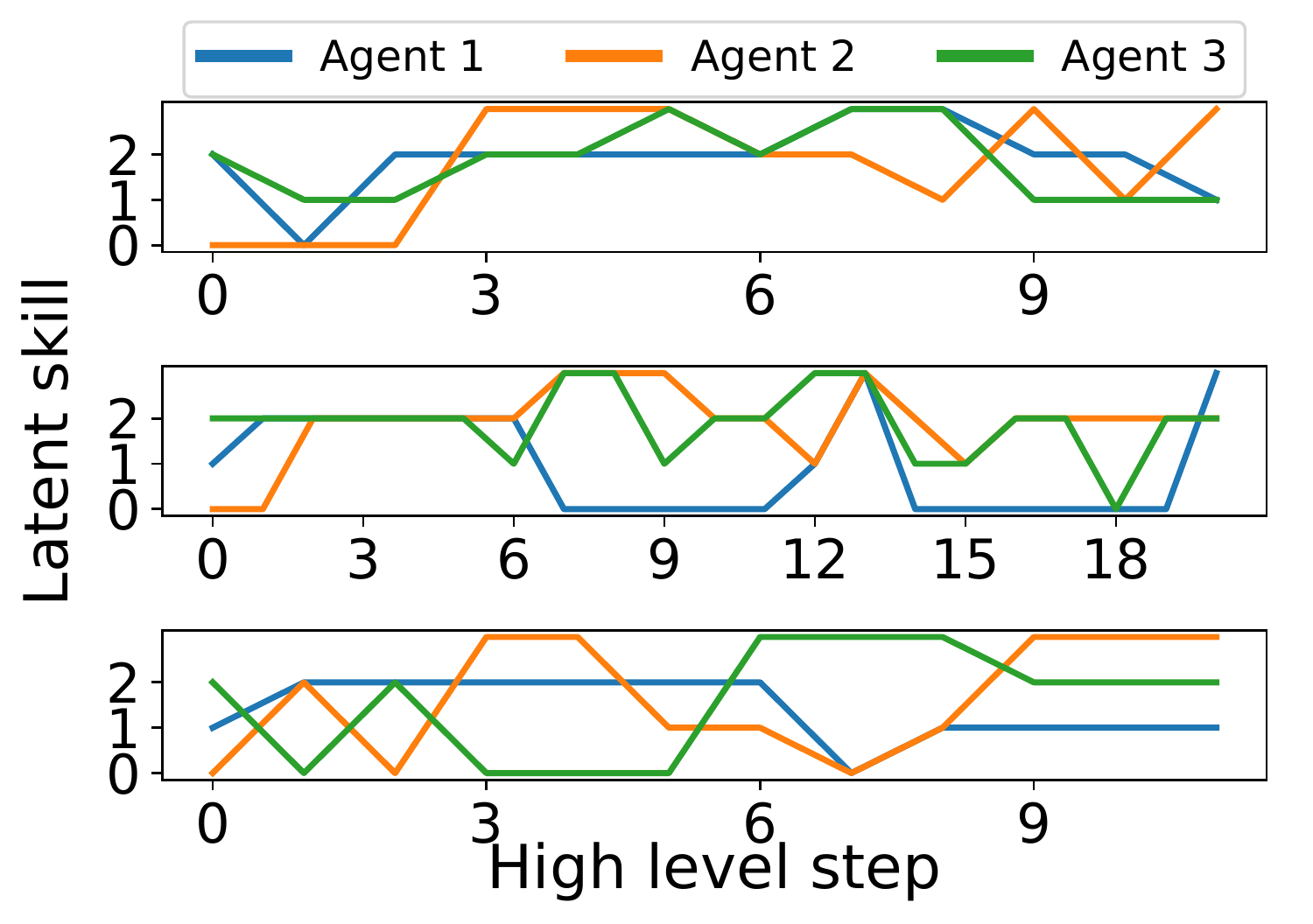}
    \caption{Time series of latent skill usage}
    \label{fig:ts-4}
\end{subfigure}
\begin{subfigure}[t]{0.64\linewidth}
    \centering
    \includegraphics[width=\linewidth]{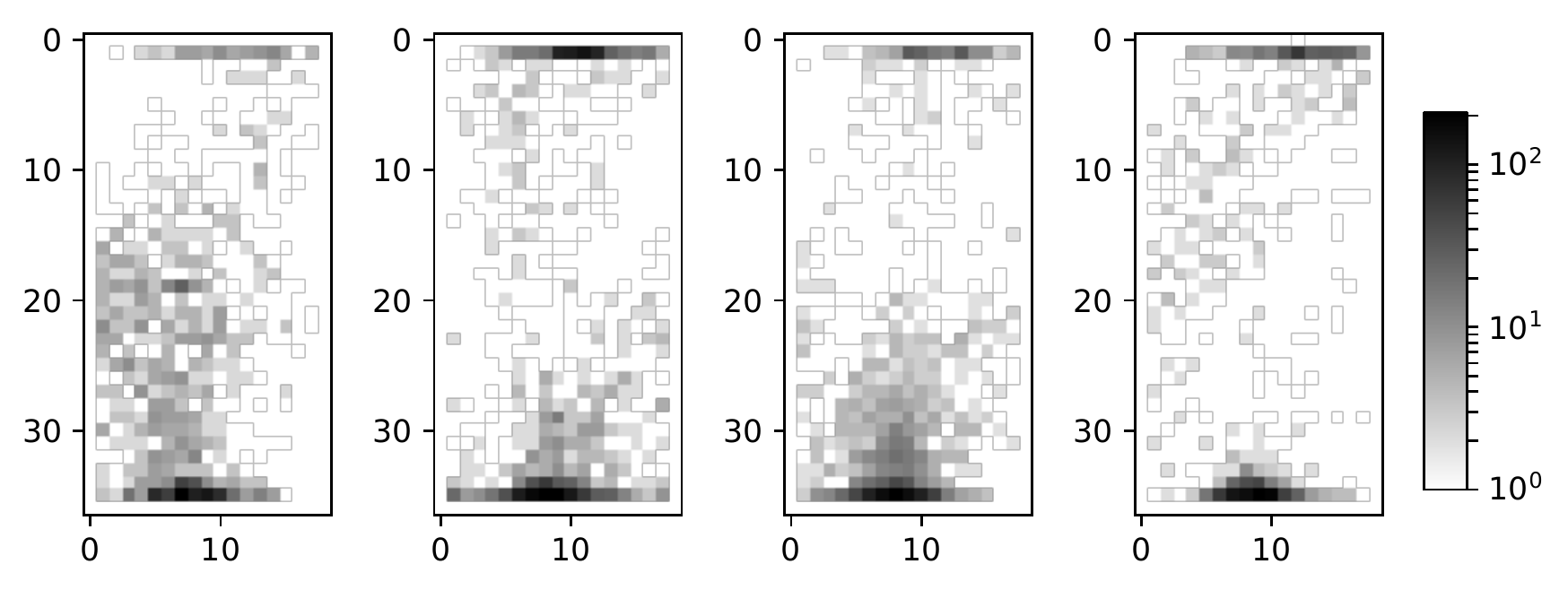}
    \caption{Latent skill usage (skills 0-3, left to right) over the spatial playing field}
    \label{fig:heatmap-4}
\end{subfigure}

\caption{(a-c): Behavioral investigation of one HSD policy, showing average and standard error over 100 test episodes. (a) Distribution of special game events for each latent skill. (b) Projection of each skill's event distribution via PCA. (c) Distribution of primitive actions for each latent skill, where ``Px'' denotes ``pass to teammate x''. (d) Projection of each skill's action distribution via PCA. (e) Count of overall skill usage, when agent team has or does not have possession. (f) Time series of skills selected high-level steps, each consisting of $t_{\text{seg}}=10$ primitive steps; each subplot shows one independent test episode; (g) Count of skill usage over the full continuous playing field, discretized to a 36x18 grid.
}
\label{fig:behavior-4}
\end{figure*}

\section{Results}
\label{sec:results}

Our method for Hierarchical learning with Skill Discovery, labeled ``HSD'', learns interpretable skills that are useful for high-level cooperation.
HSD meets the performance of QMIX and IQL, exceeds them in ad-hoc cooperation, and enables deeper policy analysis due to its hierarchical structure.
\Cref{subsec:behavior} provides a detailed quantitative behavioral analysis of learned skills.
\Cref{subsec:performance} discusses performance, hyperparameters sensitivity, and ad-hoc cooperation.

\subsection{Quantitative behavioral analysis}
\label{subsec:behavior}

\begin{figure*}[t]
\begin{subfigure}{.24\linewidth}
    \centering
    \includegraphics[width=\linewidth]{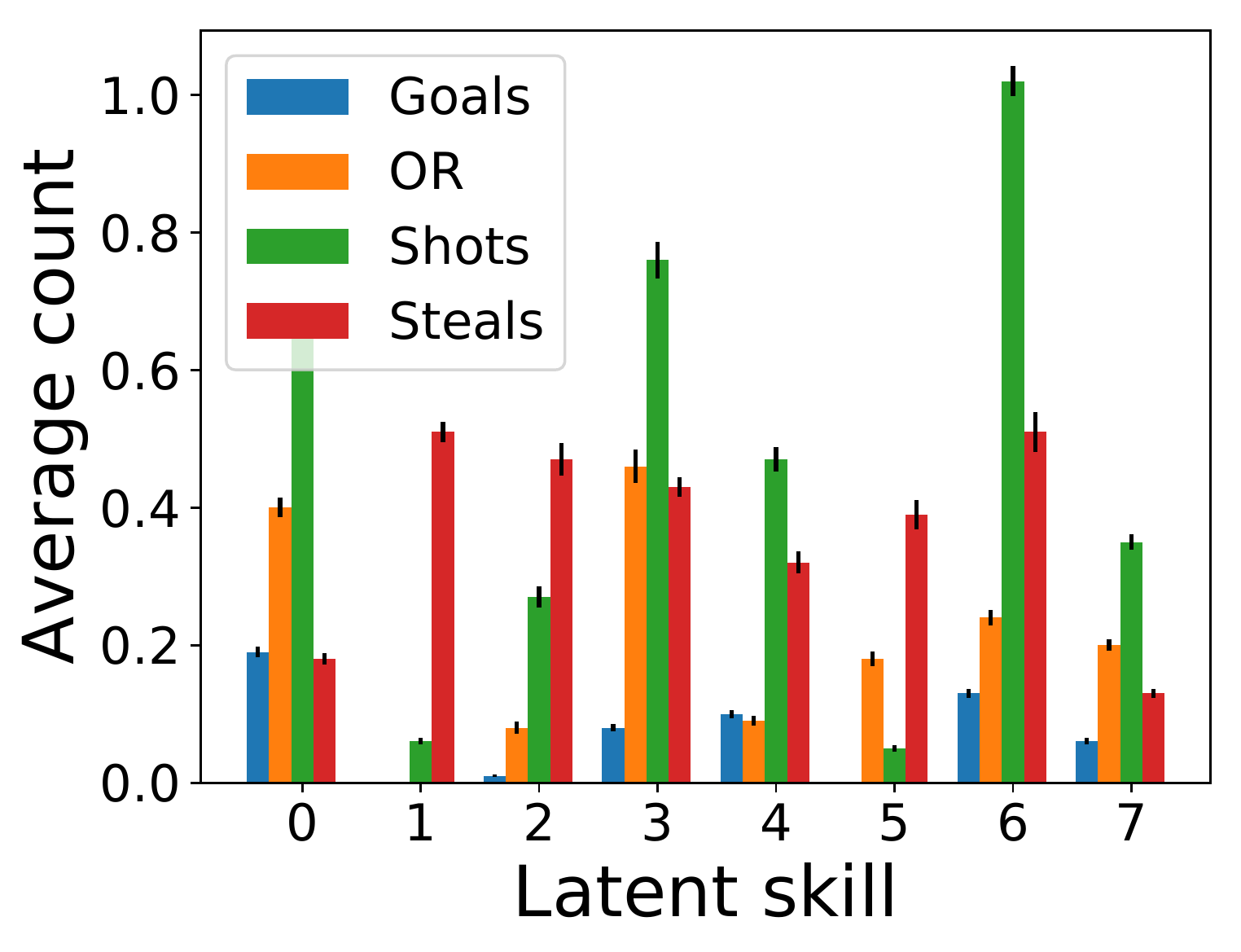}
    \caption{Event distribution}
    \label{fig:events-8}
\end{subfigure}
\begin{subfigure}{0.24\linewidth}
  \centering
  \includegraphics[width=1.0\linewidth]{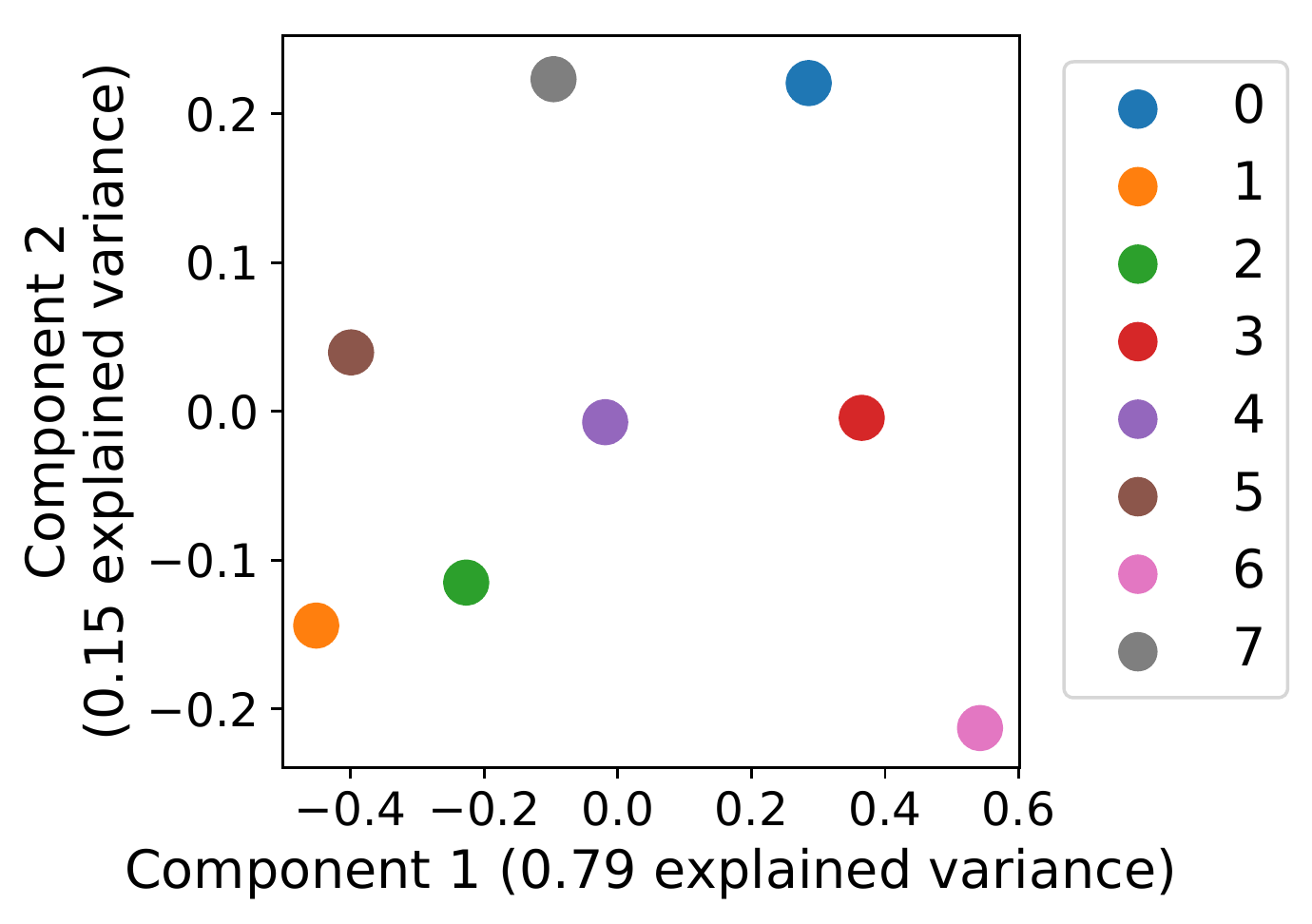}
  \caption{PCA of events}
  \label{fig:pca-events-8}
\end{subfigure}
\begin{subfigure}{0.24\linewidth}
  \centering
  \includegraphics[width=1.0\linewidth]{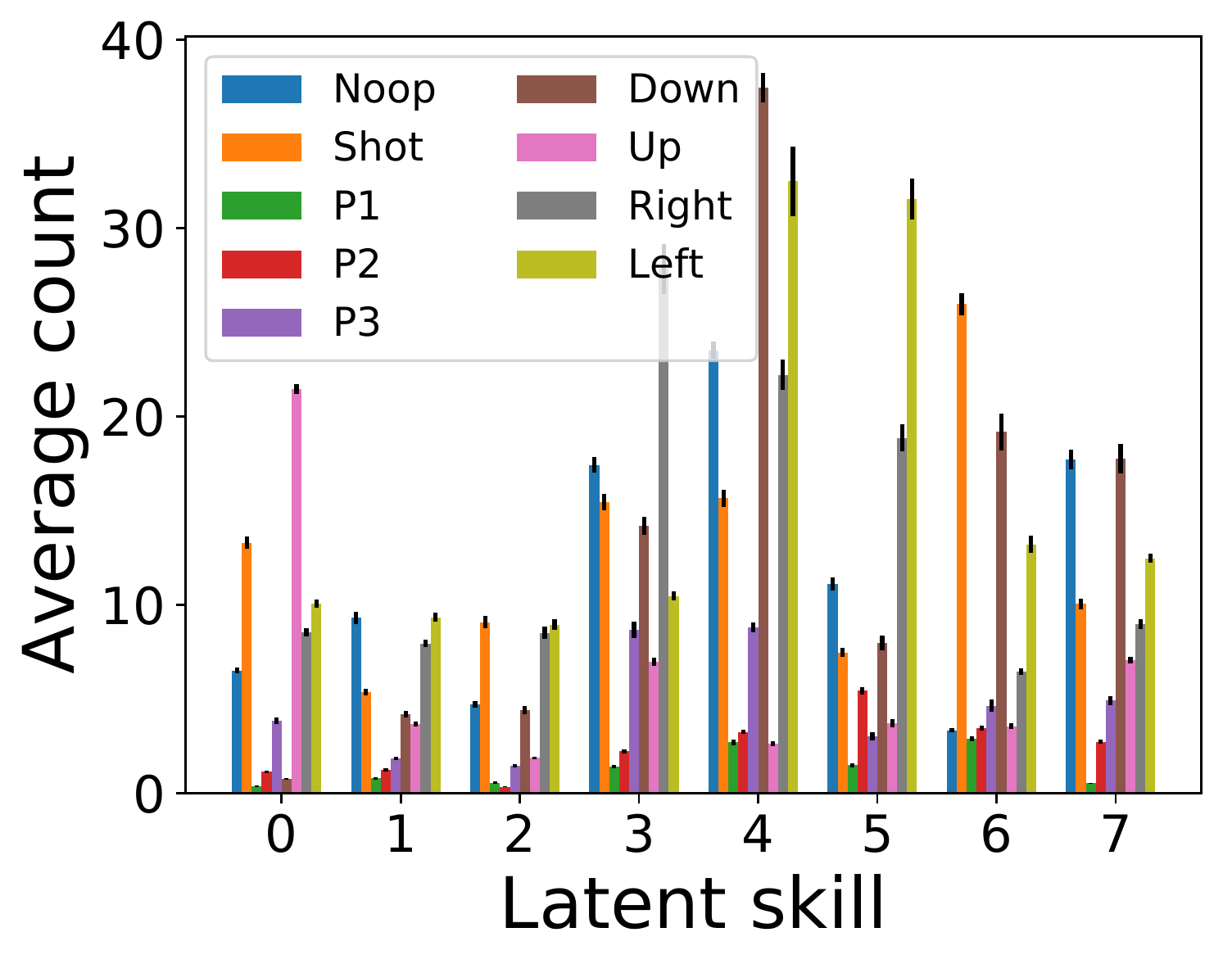}
  \caption{Action distribution}
  \label{fig:actions-8}
\end{subfigure}
\begin{subfigure}{0.24\linewidth}
  \centering
  \includegraphics[width=1.0\linewidth]{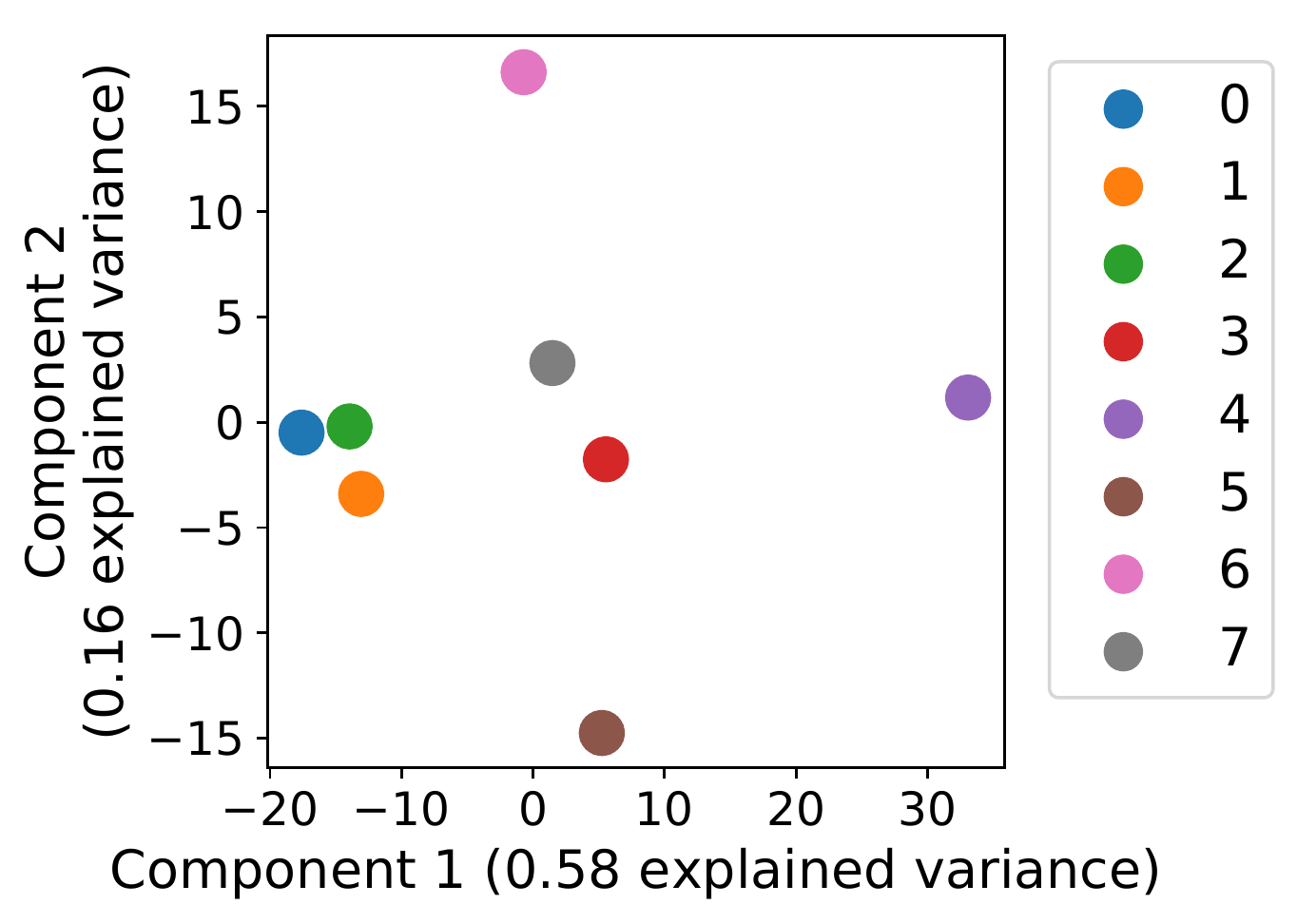}
  \caption{PCA of actions}
  \label{fig:pca-actions-8}
\end{subfigure}
\caption{Behavioral analysis of HSD policies with 8 latent skills. (a) Skill 0 makes the most goals, skill 1 focuses on defensive steals, skill 6 makes the most shot attempts. (b) Differences between skills, especially skills 0, 1 and 6, are reflected by the PCA reduction of events. (c) Skill 0 predominantly moves up, which explains its high goal rate, while skill 4 moves down the most (d) These distinguishable characteristics of skills are reflected by their large separation after PCA reduction.}
\label{fig:behavior-8}
\end{figure*}

We conducted a quantitative analysis of the discovered skills by measuring the impact of skills on occurrence of game events and primitive actions, agents' choices of skills over an episode, and the spatial occurrence of skills.
\Cref{fig:behavior-4} shows results for the case of four latent skills, which we describe immediately below.
We describe the case of eight latent skills later in \Cref{fig:behavior-8}.

\textbf{Analysis of game events. }
\Cref{fig:events-4} shows the counts of each game event under each skill, summed over any agent who was assigned to execute the skill, and averaged over 100 test episodes.
Skill 1 makes the most shot attempts, Skill 2 provides defense by focusing on steals, while Skill 3 contributes to the most number of successful goals.
This difference in game impact, which emerged without any skill-specific reward functions, is also reflected by the large separation of principal components in \Cref{fig:PCA-events-4} that result from applying PCA to the vector of event counts of \Cref{fig:events-4}.
\Cref{fig:PCA-events-4} suggests that component 1 corresponds to tendency to make offensive shots, while component 2 corresponds to tendency to make steals.
\Cref{fig:actions-4} shows the distribution of primitive actions taken by the low-level policy when conditioned on each latent skill.
Skill 0 predominantly moves up towards the opponent net to begin offense, Skill 1 is more biased toward the left field, while Skill 2 moves down to defend the home net more than other skills.
\Cref{fig:skills-4} shows the usage of each skill by the high-level policy, under the cases when agent team has possession and when the opponent team has possession.
Skill 2 is strongly associated with lack of possession since it is a defensive skill for regaining possession.

\textbf{Time series of skill usage. }
\Cref{fig:ts-4} shows a time series of skill usage over high-level steps by each agent during three different episodes (from top to bottom).
Importantly, agents learned to choose complementary skills, such as in Episode 3 when Agent 3 stays for defense while Agents 1 and 2 execute offense via Skills 1 and 3, at step 9.
Each individual agent also dynamically switches between skills, such as in Episode 1 when Agents 1 and 3 switch from the defensive Skill 2 to the offensive Skill 3 at step 6.
As shown by the extended periods in all episodes when all agents play the defensive Skill 2, agents are able to sustain the same skill over multiple consecutive high-level steps, which mitigates the concern over choosing a fixed $t_{\text{seg}}$.
Note that at any given time in the game, the defensive Skill 2 is almost always used by some agent either to make steals or cover the home net.


\textbf{Spatial occurrence of skills. }
\Cref{fig:heatmap-4} is a heatmap of skill usage over the playing field.
Consistent with the previous analysis, Skill 0 is used for moving up for offense, Skills 1 and 3 tend to camp near the opponent net (top) to attempt shots, while Skill 2 is concentrated near the home net (bottom) to make defensive steals. 

\textbf{Increasing number of skills. }
The number of latent skills is also a key design choice to make based on domain knowledge.
\Cref{fig:behavior-8} analyzes HSD when trained with eight skills.
Skills 0, 3, and 6 focus on shot attempts and offensive rebounds (\Cref{fig:events-8}), and they have high values of the first principal component (\Cref{fig:pca-events-8}).
Skills 1 and 2 focus on defensive steals.
\Cref{fig:actions-8} shows that Skill 0 moves up for offense the most, while Skill 4 moves down to play defense.
This is reflected by their large separation in the first principal component (\Cref{fig:pca-actions-8}).

\begin{figure*}[t]
\begin{subfigure}{.24\linewidth}
    \centering
    \includegraphics[width=\linewidth]{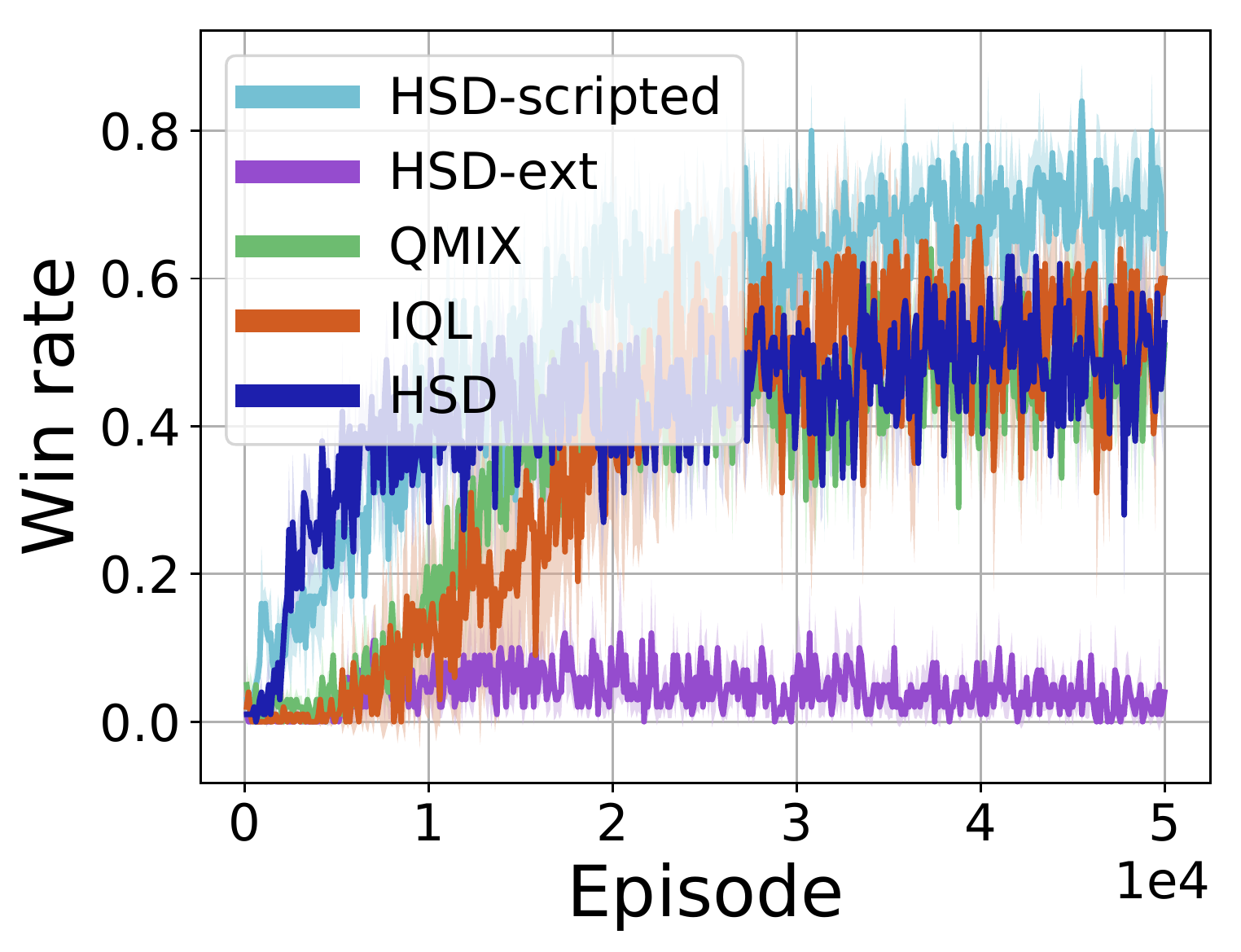}
    \caption{Win rate against scripted bots}
    \label{fig:winrate-compare}
\end{subfigure}
\begin{subfigure}{0.24\linewidth}
  \centering
  \includegraphics[width=1.0\linewidth]{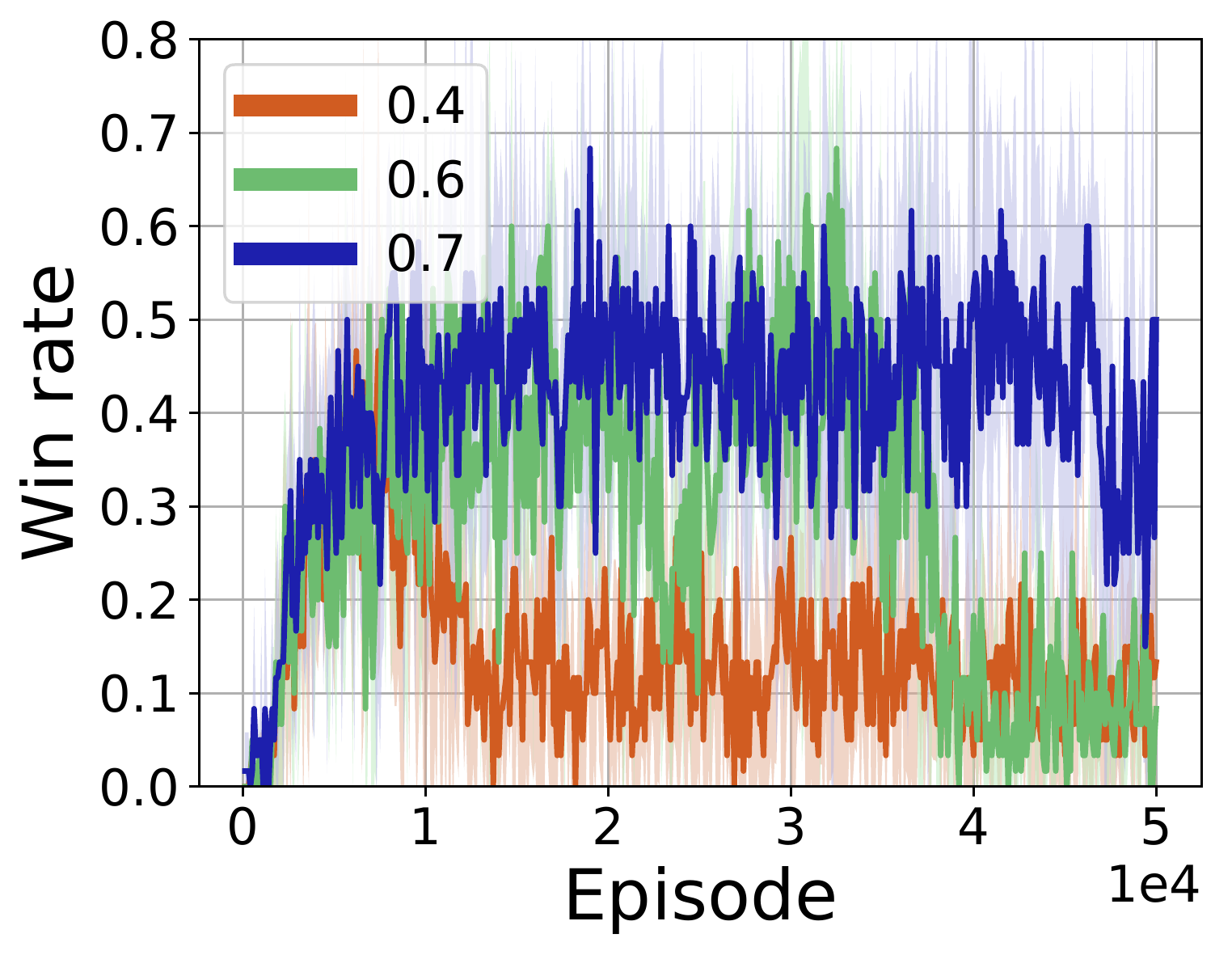}
  \caption{Varying $\alpha_{\text{threshold}}$}
  \label{fig:winrate-alphathreshold}
\end{subfigure}
\begin{subfigure}{0.24\linewidth}
  \centering
  \includegraphics[width=1.0\linewidth]{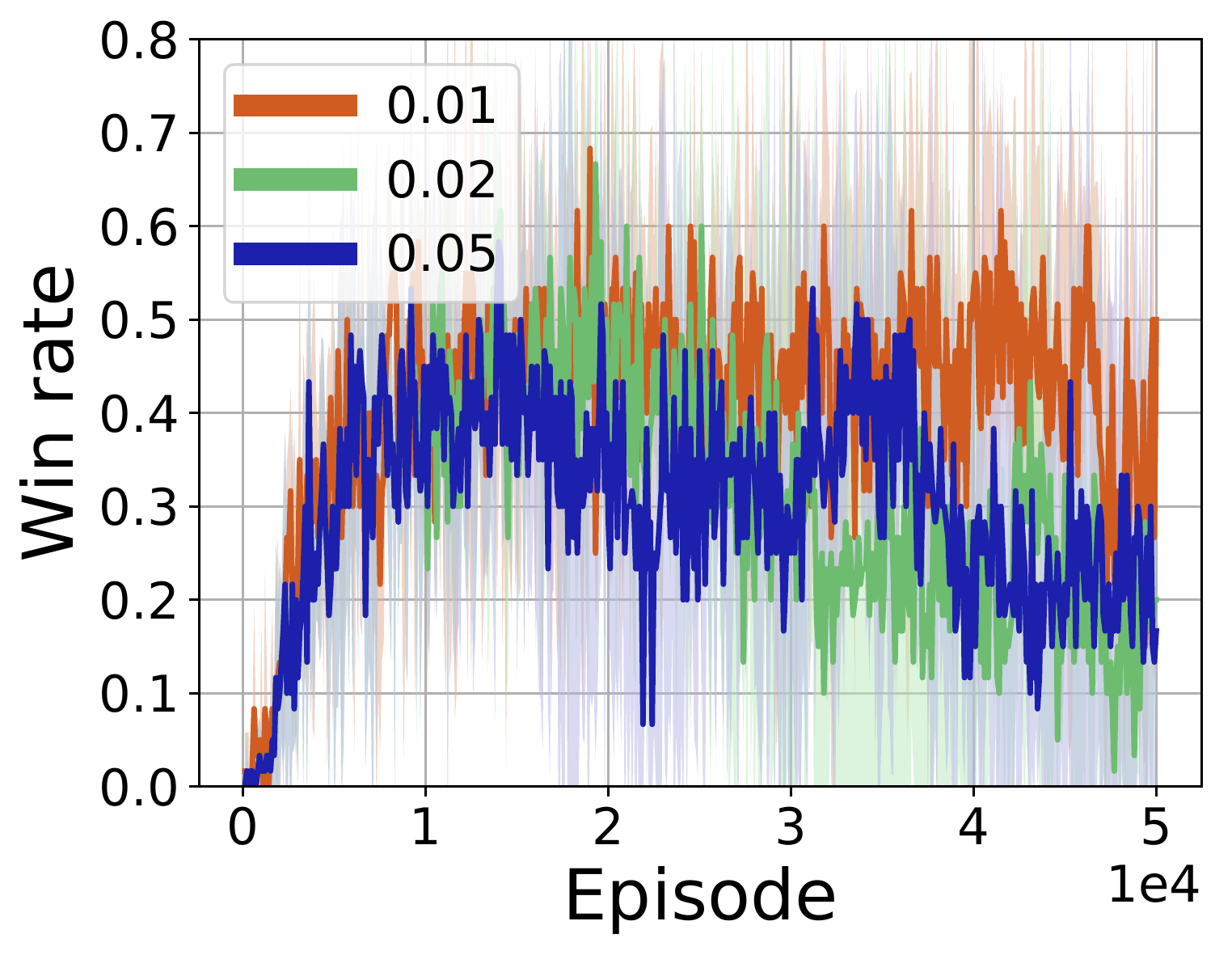}
  \caption{Varying $\alpha_{\text{step}}$}
  \label{fig:winrate-alphastep}
\end{subfigure}
\begin{subfigure}{0.24\linewidth}
  \centering
  \includegraphics[width=1.0\linewidth]{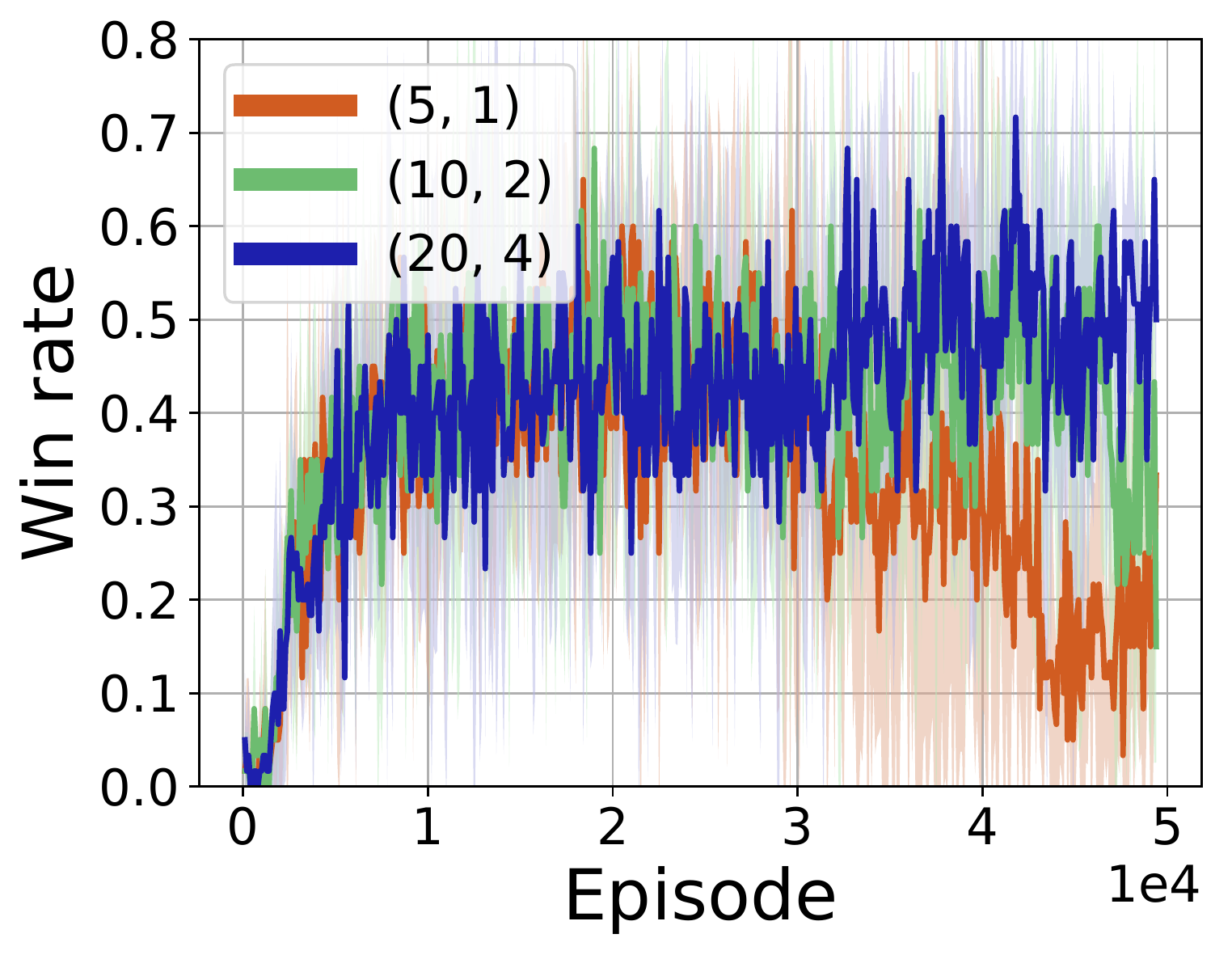}
  \caption{Varying $t_{\text{seg}}$ and $k_{\text{skip}}$}
  \label{fig:winrate-tseg}
\end{subfigure}
\caption{Win rate against scripted opponent team over training episodes.
Each curve is the mean over random seeds (5 for (a) and 3 for (b-d)) with shaded region representing 95\% confidence interval. 
(a) HSD is within margin of error with QMIX and IQL. HSD-scripted has the same hierarchical architecture as HSD but is trained with hand-scripted subtask rewards. HSD-ext does not use extrinsic rewards.
(b-d) Learning is more stable with high $\alpha_{\text{threshold}}$, small $\alpha_{\text{step}}$, and longer $t_{\text{seg}}$.}
\label{fig:winrate}
\end{figure*}

\subsection{Performance and parameter sensitivity}
\label{subsec:performance}

\Cref{fig:winrate} shows win rate against the scripted opponent team over training episodes for HSD and baselines, each with 5 independent runs, and for varying hyperparameter settings of HSD, each with 3 independent runs.
HSD agents learn faster than QMIX and IQL, consistent with findings on hierarchical versus non-hierarchical methods in early work \citep{ghavamzadeh2006hierarchical}, while their final performance are within the margin of error (\Cref{fig:winrate-compare}).
HSD-ext does not have access to extrinsic rewards and underperforms the rest.
This supports our hypothesis that the extrinsic team reward is needed in combination with the intrinsic reward to promote useful behavior.
HSD-scripted outperformed other methods, showing that using cooperative learning at the high-level and independent learning at the low level is a strong approach, and improvement to skill discovery is possible.

We investigated the effect of varying the key hyperparameters of HSD.
\Cref{fig:winrate-alphathreshold} shows that larger values of $\alpha_{\text{threshold}}$ gives higher performance and lower variance.
A small $\alpha_{\text{threshold}}$ increases the likelihood that a spuriously high evaluation performance crosses $\alpha_{\text{threshold}}$, which would cause a re-weighting of the extrinsic versus intrinsic reward even when the agents have not yet adapted to the current reward.
This explains the instability of $\alpha_{\text{threshold}}=0.4$ in \Cref{fig:winrate-alphathreshold}.
Likewise, \Cref{fig:winrate-alphastep} shows that a smaller value of $\alpha_{\text{step}}$ performs better, because each adjustment of the low-level reward is smaller and hence the automatic curriculum is easier for learning.
\Cref{fig:winrate-tseg} shows that agents who sustain high-level skills for 10 or 20 time steps perform better than agents who sustain only for 5 steps.
A smaller $t_{\text{seg}}$ means that agents make more frequent decisions to sustain or switch their choice of skill, which allows for more flexible policies but increases the difficulty of learning.

\begin{table}[h]
\caption{Win/lose percentage of final policies over 100 test episodes and 5 seeds, matched with different teammates.}
\label{table:adhoc}
    \centering
    \begin{tabular}{lrrrrrr}
    \toprule
    &
    \multicolumn{2}{c}{HSD} &
    \multicolumn{2}{c}{QMIX} & 
    \multicolumn{2}{c}{IQL} \\
    \cmidrule(r){2-3}
    \cmidrule(r){4-5}
    \cmidrule(r){6-7}
	Teammate & Win & Lose & Win & Lose & Win & Lose \\
    \midrule
    Training & 46 (4) & 39 (4) & 55 (3) & 23 (3) & 36 (7) & 46 (4) \\
    1 scripted & 49 (4) & 45 (3) & 48 (4) & 44 (4) & 32 (3) & 54 (4) \\
    2 scripted & 52 (3) & 45 (1) & 45 (2) & 51 (2) & 37 (2) & 58 (1) \\
    1 defensive & 43 (5) & 42 (4) & - & - & - & - \\
    1 offensive & 45 (2) & 41 (1) & - & - & - & - \\
    \bottomrule
    \end{tabular}
    \vspace{-10pt}
\end{table}

\textbf{Ad Hoc cooperation. }
We investigated the test performance of agents in ad-hoc cooperation, by giving them teammate(s) with whom they never previously trained \citep{stone2010ad}.
This mimics the setting where AI agents must cooperate with a human player in team sports games.
\Cref{table:adhoc} shows the win and lose percentage of HSD, QMIX, and IQL (draws are possible).
HSD agents perform as well or better when one or two of their teammates are replaced by scripted bots, possibly due to independently-trained low-level policies in HSD.
However, QMIX agents performed significantly worse when paired with scripted bots, likely because the out-of-training behavior of bots pose difficulties for QMIX agents who underwent fully-centralized training.
IQL agents also lost significantly more often with scripted teammates.
For HSD, we can also fix one agent to always play a defensive or offensive skill.
Based on \Cref{fig:events-4}, we chose Skill 1 for offense and Skill 2 for defense.
HSD agents are able to maintain their performance within the margin of error.

\section{Conclusion and discussion}
\label{sec:discussion}

We presented a method for hierarchical multi-agent reinforcement learning that discovers useful skills for strategic teamwork.
We train cooperative decentralized policies for high-level skill selection and train independent low-level policies to execute chosen skills, which emerge from a dynamically weighted combination of intrinsic and extrinsic rewards.
We demonstrated the emergence of quantifiable, distinct and useful skills in stochastic team sports simulations without assigning a reward to each skill.
These findings are a step toward multi-agent game AI that execute realistic high-level strategies and can cooperate with human players.

There are many interesting avenues for future work.
One may condition high-level policies on unique agent features, such that agents play different roles \citep{wilson2010bayesian} that affect their choice of skills.
Asynchronous termination \citep{bacon2017option} of subtasks allows learning a larger space of policies.
Optimizing the number of skills is also a natural generalization.
One may apply curriculum-learning approaches that initialize skill-conditioned low-level policies from pretraining in an induced single-agent setting \citep{yang2018cm3} or using expert data, analogous to professional players practicing skills outside of team matches.
This may speed up training since low-level policies can already generate useful trajectories that can be segmented into distinguishable skills.

\begin{acks}
We are grateful to Ahmad Beirami (Facebook AI) for significant contributions to extensive and insightful discussions throughout the course of this work, and for detailed feedback that helped to improve the clarity and precision of the paper.
We also thank individuals from Electronic Arts: Maziar Sanjabi and Yunqi Zhao for discussions; Caedmon Somers, Jason Rupert and Yunqi Zhao for STS2 development; Mohsen Sardari and Kazi Zaman for support.
\end{acks}


\bibliographystyle{ACM-Reference-Format}  
\bibliography{citation}  

\end{document}